%% file: arxiv.tex
\documentclass{article}

\usepackage{arxiv}

\usepackage{amsmath}
\usepackage{amssymb}
\usepackage{amsthm}
\usepackage{graphicx}

\usepackage[algoruled,vlined,linesnumbered]{algorithm2e}
\usepackage{algorithmic}

\usepackage{booktabs}
\usepackage{multirow}
\usepackage{siunitx}
\usepackage{array}

\usepackage{longtable}
\usepackage{url}
\usepackage{hyperref}
\usepackage{natbib}

\usepackage{svg}

\allowdisplaybreaks

\title{A Heuristic Algorithm Based on Beam Search and Iterated Local Search for the Maritime Inventory Routing Problem}

\author{
\href{https://orcid.org/XXX}{\includegraphics[scale=0.1]{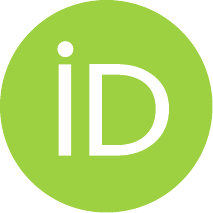}\hspace{1mm}Nathalie Sanghikian} \\
Departamento de Engenharia Industrial \\
Pontif\'{\i}cia Universidade Cat\'olica do Rio de Janeiro \\
\texttt{nat\_sanghikian@aluno.puc-rio.br} \\
\And
\href{https://orcid.org/XXX}{\includegraphics[scale=0.1]{figures/orcid.pdf}\hspace{1mm}Rafael Meirelles} \\
Departamento de Engenharia Industrial \\
Pontif\'{\i}cia Universidade Cat\'olica do Rio de Janeiro \\
\texttt{rafaelpmeirelles@aluno.puc-rio.br} \\
\And
\href{https://orcid.org/0000-0001-5715-149X}{\includegraphics[scale=0.1]{figures/orcid.pdf}\hspace{1mm}Rafael Martinelli} \\
Departamento de Engenharia Industrial \\
Pontif\'{\i}cia Universidade Cat\'olica do Rio de Janeiro \\
\texttt{martinelli@puc-rio.br} \\
\And
\href{https://orcid.org/XXX}{\includegraphics[scale=0.1]{figures/orcid.pdf}\hspace{1mm}Anand Subramanian} \\
Departamento de Informática \\
Universidade Federal da Paraíba \\
\texttt{anand@ci.ufpb.br}
}

\date{June 4th, 2025}

\renewcommand{\headeright}{A Preprint}
\renewcommand{\undertitle}{A Preprint}
\renewcommand{\shorttitle}{A Heuristic Algorithm Based on Beam Search and Iterated Local Search for the MIRP}

\hypersetup{
pdftitle={A Heuristic Algorithm Based on Beam Search and Iterated Local Search for the Maritime Inventory Routing Problem},
pdfsubject={math.OC},
pdfauthor={Nathalie Sanghikian, Rafael Meirelles, Rafael Martinelli, Anand Subramanian},
pdfkeywords={Maritime Inventory Routing Problem, Heuristics, Beam Search, Iterated Local Search},
}

\begin{document}
\maketitle

\begin{abstract}
\input{sections/00.Abstract}
\end{abstract}



\input{sections/01.Introduction}
\input{sections/02.LiteratureReview}
\input{sections/03.ProblemDescription}
\input{sections/04.ProposedHeuristic}
\input{sections/05.ComputationalExperiments}
\input{sections/06.Conclusions}
\input{sections/07.Acknowledgements}

\bibliographystyle{elsarticle-harv}
{\linespread{1.3}\selectfont\bibliography{cas-refs}}

\appendix

\input{sections/AppendixA}
\input{sections/AppendixB}

\end{document}

%% file: sections/00.Abstract.tex
Maritime Inventory Routing Problem (MIRP) plays a crucial role in the integration of global maritime commerce levels. However, there are still no well-established methodologies capable of efficiently solving large MIRP instances or their variants due to the high complexity of the problem. The adoption of exact methods, typically based on Mixed Integer Programming (MIP), for daily operations is nearly impractical due to the CPU time required, as planning must be executed multiple times while ensuring high-quality results within acceptable time limits. Non-MIP-based heuristics are less frequently applied due to the highly constrained nature of the problem, which makes even the construction of an effective initial solution challenging. Papageorgiou et al. (2014) introduced a single-product MIRP as the foundation for MIRPLib, aiming to provide a collection of publicly available benchmark instances. However, only a few studies that propose new methodologies have been published since then. To encourage the use of MIRPLib and facilitate result comparisons, this study presents a heuristic approach that does not rely on mathematical optimization techniques to solve a deterministic, finite-horizon, single-product MIRP. The proposed heuristic combines a variation of a Beam Search algorithm with an Iterated Local Search procedure. Among the 72 instances tested, the developed methodology can improve the best-known solution for 19 instances within an acceptable CPU time.

%% file: sections/01.Introduction.tex
\section{Introduction}
\label{sec:intro}

In the international shipping industry, operating companies often have to manage product transport within their fleet, respecting inventory levels in ports and vessels in an integrated way. The combination of these elements takes into account the route and scheduling of the ships with inventory management, known in the literature as the \emph{Maritime Inventory Routing Problem} (MIRP) \citep{christiansen2009maritime}. In this challenging combinatorial problem, the routing and scheduling steps define which ports will be visited by each vessel, in which sequence and time. Meanwhile, the inventory management step determines the inventory levels of each port and vessel at each point in time within a given planning horizon (\cite{christiansen2009maritime}, \cite{Costa2018}).  

Solving MIRPs is a challenging task due to its alignment with real-world complexities, such as variable demand rates, multi-product logistics, and the coordination of a central supplier with multiple consumers.
\cite{Fagerholt2023MaritimeIR} reviewed MIRP research from the past decade and proposed several future directions, including applications in energy transport, such as hydrogen, ammonia, LNG, and liquefied CO2, considering increasingly aggressive decarbonization goals. 
They also emphasize a key consideration when addressing this type of problem: the different planning levels, particularly between the tactical and operational levels. Strategic and tactical planning problems typically involve decisions with long-term impacts, while operational planning focuses on planning over shorter time horizons, ranging from a few days or weeks to a few months, and tends to involve much harder constraints. In addition, operational planning problems require tools capable of providing high-quality results in acceptable time frames. As a result, exact methods are generally not favored as they typically deliver optimal solutions only for small-scale instances of practical size. Another challenge involving MIRP is the strong interdependence between routing decisions, where factors such as inventory levels and berth limitations trigger chain reactions throughout the supply chain.

Over the past two decades, matheuristics have emerged as a growing field in Operations Research. These are optimization algorithms that combine metaheuristics with mathematical programming techniques \citep{papageorgiou2018recent}, often used in the MIRP literature. On the other hand, pure heuristics (i.e. those that do not rely on mathematical optimization procedures) are less commonly applied due to the very constrained nature of the problem, as just building an initial solution can be very challenging \citep{FRISKE2022105520}.

There is no single and well-defined definition of an MIRP, but rather numerous variants tailored to specific industrial applications (\cite{papageorgiou2014mirplib}, \cite{andersson2010industrial}). Given this diversity, standardizing the comparison of solution methods and establishing a taxonomy is essential. To address this issue, \cite{papageorgiou2014mirplib} introduced the MIRPLib, a collection of publicly available single-product MIRP benchmark instances. A subset of these instances, called Group 2, serves as the focus of this study. These instances feature a planning horizon of over 120 periods, with no split pickups or deliveries. The problem setting includes a single production port, multiple consumer ports, and various classes of vessels.

In this work, our objective is to develop a pure heuristic approach for MIRP applied to instances of Group 2 by \cite{papageorgiou2014mirplib} available in MIRPLib. The heuristic consists of a variation of the Beam Search (BS) algorithm (\cite{lowerre1976harpy}, \cite{OW01011988}, \cite{Reddy}) with an Iterated Local Search (ILS) \citep{lourencco2003iterated} (BS-ILS).

The main contributions of this work are: (i) the introduction of an innovative solution representation for MIRP; (ii) the development of a new approach using the BS-ILS algorithm to address a deterministic MIRP within reasonable CPU times; and (iii) a Julia package designed to support the applicability of the MIRPLib.


This paper is organized as follows. Section \ref{sec:litrev} briefly reviews the literature on deterministic MIRP works and its variants, and presents heuristic applications. Section \ref{sec:probldesc} describes the deterministic MIRP considered in this paper. Section \ref{sec:simmath} presents the proposed BS-ILS heuristic. Section \ref{sec:results} reports and discusses the results of computational experiments. Finally, Section \ref{sec:conc} concludes the work and provides directions for future research.

%% file: sections/02.LiteratureReview.tex
\section{Literature Review}
\label{sec:litrev}

\subsection{Maritime Inventory Routing Problem}

The MIRP refers to the strategic coordination of maritime transportation and inventory management, integrating logistics decisions with dynamic inventory levels at both origins and destinations. It requires planning vessel routes, determining port call sequences and deciding product quantities for loading and unloading, while ensuring that inventory limits are respected \citep{sanghikian2020}. 

Real-world MIRP applications are often complex, incorporating variations such as supplier/producer or central consumer configurations, inventory constraints at specific ports, variable production/consumption rates, multiple products, and spot charters. To support the development and comparison of solution methods for these variants, standardized instances and modeling frameworks are essential. In this context, \citet{papageorgiou2014mirplib} introduced the MIRPLib to standardize MIRP definitions and support advanced formulations. They proposed a core model for single-product distribution, ensuring inventory constraints at all ports. Building on this foundation, additional MIRP variants were developed, including Group 2 instances — deterministic, long-horizon problems with one loading port and multiple discharge ports. This simplified model assumes that the vessels travel fully loaded to discharge regions and return empty. Detailed descriptions of the original Group 2 model are provided in Section \ref{sec:probldesc}.

\subsection{Heuristic Approaches}
\label{subsec:heuristics}

Many authors have developed methods to address MIRP, including mathematical models, heuristics, metaheuristics, and matheuristics. \cite{SHAABANI2023106214} and \cite{Fagerholt2023MaritimeIR} provided a comprehensive review of the literature on MIRP and the algorithms used. They summarized and classified recent MIRP studies, emphasizing factors such as the size of the problem (e.g., the number of ports, vessels, products, and planning horizon) and the techniques applied to solve them. Out of the 34 studies listed by \cite{SHAABANI2023106214} and the 53 studies listed by \cite{Fagerholt2023MaritimeIR}, 38 focus on MIRP solutions using heuristic techniques.

\cite{ronen2002marine} pioneered this area by introducing a cost-based heuristic. A decade later, \cite{staalhane2012construction} proposed a multi-start local search heuristic for the LNG inventory routing problem, using greedy insertion for initialization and refining solutions via neighborhood search, branch-and-bound, or their combination. \cite{uggen2013using} implemented a fix-and-relax time decomposition heuristic for MIRPs with a single product, balancing solution quality with computational efficiency, and \cite{goel2015constraint} proposed an iterative heuristic for LNG ship scheduling and inventory management. 

Metaheuristic approaches have become important for exploring large search spaces. \cite{christiansen2011maritime} integrated a construction heuristic with a Genetic Algorithm (GA) for multi-product MIRPs. \cite{siswanto2011multi} proposed a multi-heuristic methodology that achieved optimality for smaller instances. Expanding further, \cite{siswanto2011solving} developed a multi-heuristic strategy for a ship inventory routing and scheduling problem with non-dedicated compartments. More recently, \cite{siswanto2019maritime} proposed a GA based on multi-heuristics for an MIRP with multiple time windows. \cite{sheikhtajian2020marine} studied uncertain LNG marine IRPs and proposed a metaheuristic that combines heuristics, GA, and fuzzy modeling to improve sensitivity analysis.

Hybrid metaheuristics have also been explored. \cite{chandra2013multi} combined Ant Colony Optimization with linear programming for MIPRs with optional cargoes, while \cite{song2013maritime} applied a large neighborhood search (LNS) heuristic to iteratively solve subproblems. \cite{de2017sustainable} applied Particle Swarm Optimization (PSO) to sustainable MIRPs with time windows, while \cite{zhang2018flexible} used a Lagrangian heuristic to manage delivery time uncertainty. \cite{al2022optimization} investigated a short-term LNG scheduling problem, analyzing transportation expenses and emissions using PSO.

Matheuristics, which integrate mathematical programming and heuristics, are a powerful tool for solving MIRPs. \cite{agra2014hybrid} introduced hybrid heuristics for short-sea inventory routing, combining rolling-horizon heuristics (RHH) with local branching and feasibility pumps. \cite{shao2015hybrid} proposed a two-stage heuristic for LNG-IRPs, integrating a RHH with MIP-based neighborhood refinement methods. \cite{hemmati2015effective} and \cite{hemmati2016iterative} developed hybrid matheuristics for cargo and inventory routing in tramp shipping. Meanwhile, \cite{mutlu2016comprehensive} focused on the LNG Annual Delivery Program (LNG-ADP), which is a schedule of LNG deliveries for a given year, introducing a matheuristic with a unique vessel selection methodology.

\cite{agra2016mip} developed a MIP-based heuristic with local search techniques, while \cite{agra2017combined} proposed a matheuristic to improve solutions and ensure feasibility when the mathematical model fails. \cite{agra2018robust} and \cite{diz2019robust} refined MIRP decomposition algorithms using an Iterated Local Search (ILS) heuristic and robust optimization. \cite{sanghikian2020} designed a variable neighborhood search (VNS) for a multi-product MIRP, integrating LP solvers to optimize continuous variables for routing solutions.

Recent research has focused on hybrid methods, parallel strategies, and uncertainty modeling. \cite{dong2018reoptimization} developed a RHH re-optimization framework for combined maritime fleet deployment and inventory management with port-visit flexibility in roll-on/roll-off shipping. \cite{stanzani2018optimizing} proposed a multi-start heuristic for the simultaneous pickup and delivery ship routing, incorporating two enhancement methods and a local search for further solution improvement. Furthermore, \cite{msakni2018short} applied an optimization-based VNS for short-term LNG delivery planning, using a greedy heuristic for initialization and iteratively solving reduced instances with a commercial solver. For LNG-ADPs problems, \cite{li2020planning} proposed an RHH, while \cite{li2023maritime} incorporated transshipment and intermediate storage, using various RHH configurations. 

In MIRPLib-specific studies, \cite{papageorgiou2018recent} introduced several heuristics, such as RHH, $K$-opt heuristics, local branching, solution polishing and hybrid approaches, that combine metaheuristics with mathematical programming to solve MIRPLib's Group 2 instances. Their study resulted in new best-known solutions for 26 out of 70 instances that had not yet been proven optimal, and they achieved new best-known lower bounds for 56 instances. \cite{munguia2019tailoring} presented the Parallel Alternating Criteria Search, an LNS-based heuristic that yielded strong results for Groups 1 and 2.

\cite{friske2020multi} implemented a multi-start heuristic with LNS for Group 1 instances, while \cite{FRISKE2022105520} employed a relax-and-fix algorithm followed by fix-and-optimize to refine MIRPLib's Group 1 solutions. \cite{wang2023maritime} introduced a bilevel parallel memetic algorithm for a maritime location inventory routing problem, tested in Group 1 instances, achieving near-best solutions with favorable deviations and faster computation times.

In contrast to the reviewed studies, which often rely on mathematical models, this study employs a heuristic approach that avoids mathematical programming techniques. This method minimizes the dependency on complex decision models while maintaining practical solution quality, distinguishing itself through accessibility and applicability to real-world scenarios.

%% file: sections/03.ProblemDescription.tex
\section{Problem Description}
\label{sec:probldesc}

The formulation presented in this section corresponds to the MIP model for Group 2 instances available in MIRPLib \citep{papa2013}. The Group 2 instances describe long-horizon deterministic single product maritime inventory routing problems. In the following description, the terms of the sets ``port'' ($J$) and ``region'' ($R$) are used as synonyms, since it is assumed that there is one port per region.

In a defined time horizon $T$, a heterogeneous fleet of ships, classified into different vessel classes $VC$, each with distinct capacities $Q^{vc}$, is assigned to service the production ports $J^P$ and the consumption ports $J^C$. This operation must be done while ensuring that the minimum and maximum inventory limits in port $j \in J$ ($S_{j,t}^{min}$ and $S_{j,t}^{max}$, respectively) are respected. Each port $j \in J$ is characterized by an initial inventory $s_{j,0}$ and a specific production or consumption rate $d_{j,t}$ in period $t \in T$, along with an associated indicator parameter $\Delta_j$, which takes the value $1$ for production ports and $-1$ for consumption ports. Berth limit ($B_j$) constraints must also be respected, restricting the number of vessels that can attempt to load or discharge in a port $j \in J$ in a given period of time. In this MIRP, a vessel must travel at full capacity from a loading port to a discharging port, and empty from a discharging port to a loading port.

The formulation follows an arc-flow mixed-integer linear programming model, where a set of nodes $\mathcal{N}$ represents the port-time pairs. A set of arcs $A^{vc}$ associated with each vessel class $vc \in VC$ defines the possible routes between nodes. Furthermore, a set of nodes $\mathcal{N}_{s,t}$ is defined, including both the source $n_s$ and the sink $n_t$ nodes. The set of inter-regional arcs $A^{vc,ir}$ represents the arcs between the production and consumption regions. For each vessel class $vc \in VC$ and for each node $n \in \mathcal{N}_{s,t}$, the incoming and outgoing arcs are denoted by $RS^{vc}_{n}$ (for reverse star) and $FS^{vc}_{n}$ (for forward star), respectively, while the set of forward star inter-regional arcs is denoted by ($FS^{vc,ir}_{n}$).

All routing costs are incorporated into $C^{vc}_{a}$ (cost for vessel of vessel class $vc \in VC$ to traverse the arc $a \in A^{vc}$). Arc costs may vary depending on the type of arc (source, sink, waiting, inter-port). The distance $\delta_{j_1,j_2}$ between ports $j_1$ and $j_2$, the cost per kilometer per vessel $C^{vc}_{km}$ according to its vessel class $vc \in VC$ and the fees $\pi_j$ of each port $j \in J$ are considered. Some discounts are also included, a reward for finishing the route before the time horizon ends $\rho$ and a discount $\gamma^{vc}$ for vessels of a vessel class $vc \in VC$ traveling empty from a consumption port to a production port. Inventory costs $P_{j,t}$ are included as penalties, reflecting production and consumption in each port $j \in J$ to comply with inventory constraints in each period $t \in T$.

Integer decision variables $x_a^{vc}$ represent the number of vessels in a given vessel class $vc \in VC$ that traverse a specific arc $a \in A^{vc}$. The continuous decision variables $\alpha_{j,t}$, representing a spot charter, represent the amount to be produced or consumed in port $j \in J$ during period $t \in T$, and $s_{j,t}$ the quantity of inventory units in port $j \in J$ at the end of period $t \in T$.

The MIRP model \citep{papa2013} is described as follows:

\noindent 
\begin{flalign}
\label{eq:FO}
\max z = 
& \sum\limits_{vc \in VC}\sum\limits_{a \in A^{vc}} - C_a^{vc} x_a^{vc} 
&& \text{(routing cost)} \nonumber \\ 
& + \sum\limits_{j \in J}\sum\limits_{t \in T} - P_{j,t} \alpha_{j,t} 
&& \text{(inventory penalties cost)} \phantom{space}
\end{flalign}

\noindent subject to

{\begingroup
\allowdisplaybreaks 
\begin{flalign}
\label{eq:cons_MP-1}
& \sum\limits_{a \in FS_n^{vc}} x_a^{vc} - \sum\limits_{a \in RS_n^{vc}} x_a^{vc} =
\begin{cases}
1 & \text{if } n = n_s, \\
-1 & \text{if } n = n_t, \\
0 & \text{if } n \in \mathcal{N} \setminus \{n_s, n_t\}
\end{cases}
&&  n \in \mathcal{N}_{s,t}, \, vc \in VC \\[6pt]
\label{eq:cons_MP-2}
& s_{j,t} = s_{j,t-1} + \Delta_j \left( d_{j,t} - \sum\limits_{vc \in VC}\sum\limits_{a \in FS_n^{vc,ir}} Q^{vc} x_a^{vc} - \alpha_{j,t} \right)  
&&  n = (j,t) \in \mathcal{N} \\[6pt]
\label{eq:cons_MP-3}
& S_{j,t}^{\min} \leq s_{j,t} \leq S_{j,t}^{\max} 
&&  n = (j,t) \in \mathcal{N} \\[6pt]
\label{eq:cons_MP-4}
& \sum\limits_{vc \in VC}\sum\limits_{a \in FS_n^{vc,ir}} x_a^{vc} \leq B_j 
&&  n = (j,t) \in \mathcal{N} \\[6pt]
\label{eq:cons_MP-5}
& \alpha_{j,t} \geq 0 
&&  n = (j,t) \in \mathcal{N} \\[6pt]
\label{eq:cons_MP-6}
& x_a^{vc} \in \{0,1\} 
&&  vc \in VC, \, a \in A^{vc,ir} \\[6pt]
\label{eq:cons_MP-7}
& x_a^{vc} \in \mathbb{Z_+}  
&&  vc \in VC, \, a \in A^{vc} \setminus A^{vc,ir}.
\end{flalign}
\endgroup}

In this model, the objective function \eqref{eq:FO} minimizes the sum of transportation costs, and penalties for lost production and stockout. Constraints \eqref{eq:cons_MP-1} ensure the flow balance of vessels within each vessel class. The minimum and maximum inventory limits at ports must be respected, while also ensuring the continuity of inventory flow, as defined by Constraints \eqref{eq:cons_MP-2} and \eqref{eq:cons_MP-3}, respectively. Constraints \eqref{eq:cons_MP-4} represent the berth limit constraints. Constraints \eqref{eq:cons_MP-5} ensure that the spot charter variable is always non-negative. Constraints \eqref{eq:cons_MP-6} ensure that at most one vessel per vessel class travels along a particular interregional arc. Finally, Constraints \eqref{eq:cons_MP-7} define the variables' domains for the remaining arcs.

Solving MIRP using exact mathematical models is often impractical due to their high computational cost. As the problem size increases, particularly in real-world instances, solution times become unmanageable, limiting the applicability of exact approaches. Consequently, exact methods are limited to small or simplified cases, highlighting the need for scalable alternatives like heuristics or metaheuristics that can provide high-quality solutions within a reasonable CPU time.

%% file: sections/04.ProposedHeuristic.tex
\section{Proposed Heuristic}
\label{sec:simmath}

The developed heuristic is a modified version of Beam Search (\cite{lowerre1976harpy}, \cite{OW01011988}, \cite{Reddy}), which serves as a constructive algorithm, complemented by local search procedures that improve all the generated solutions. Once these are concluded, an Iterated Local Search \citep{lourencco2003iterated} is applied to possibly refine the incumbent solution. 

\begin{figure}[!htbp]
    \centering
	\includegraphics[width=0.8\textwidth]{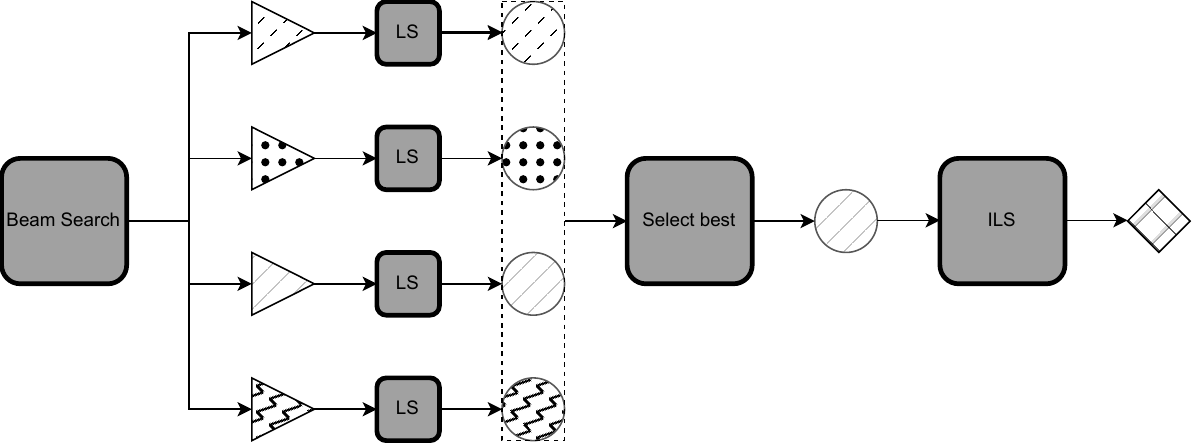}
	\caption{Proposed Heuristic.\label{fig:algorithm}}
\end{figure}

A high-level overview of the algorithm is presented in Figure \ref{fig:algorithm}. The process begins with a modified version of the BS heuristic which explores a limited set of the most promising solution paths at each step. As further detailed in Section \ref{sec: Beam Search Algorithm}, instead of expanding all possible successors, as in a traditional breadth-first search, it maintains only a fixed number of the best candidates, reducing computational effort while directing the search towards high-quality solutions, shown as triangles. These solutions undergo a local search (LS) improvement based on a Randomized Variable Neighborhood Descent (RVND) procedure \citep{Subramanianetal2010}, indicated as circles. Finally, the incumbent solution is possibly further refined by the ILS approach, yielding the final solution, presented in diamond form.

\subsection{Solution Representation}
\label{sec:Representation}



The solutions are represented by a single vector that stores port-vessel information, defined as a \textit{call}, and the order of the calls in the vector corresponds to the chronological order of the ships' visits to these ports. For a call to occur, the preceding ones must already have taken place, with respect to the port and the vessel. Thus, a ship requiring to service a call in a port must wait its turn in the list to perform the operation. The sequence is chronological, but only when it involves the same ports or vessels across the calls. For example, if the first port to be serviced is port P1 by vessel V3, followed by port P1 again, but this time by vessel V1, vessel V3 must end its operations at port P1 before vessel V1 can begin (considering that there is only one berth available at port P1). On the other hand, if the next call is unrelated, it may occur prior to the last one. To ensure feasibility, vessels must always travel between a production port and a consumption port, in either direction. This rule prevents vessels from traveling between ports of the same type, ensuring operational consistency.


This solution representation allows a given solution to have multiple representations, leading to redundancy. Two solutions can correspond to the same actual solution, as the calls involve distinct ships servicing different ports. For example, a solution may evaluate (P1,V1) followed by (P3,V2), while another solution evaluates (P3,V2) first and then (P1,V1). Despite the difference in visit order, both solutions correspond to the same operation in real life. This illustrates that permutations may not affect the final outcome and can be considered equivalent in the problem context.

Note that it is not straightforward to check which solutions are equivalent. A specific approach has been created to address this issue with some auxiliary structures as follows. For each element in the vector, four auxiliary pointers are maintained, each storing the position of the last occurrence of the vessel, the last occurrence of the port, the next occurrence of the vessel, and the next occurrence of the port. This information is preprocessed and used during local search procedures to restrict their moves to non-redundant operations, avoiding unnecessary computation. 
In addition, the time for each vessel and port is also preprocessed to speed up the solution evaluation.

\subsection{Solution Evaluation}
\label{sec: Solution Evaluation}

With the aim of evaluating a solution, each call keeps key information such as the most recent service time at the respective port, the last service time of the vessel, inventory levels, the time of the next violation and the accumulated solution costs (including both routing and inventory components), in addition to the previously described pointers. Depending on the type of change introduced, whether by local search, ILS perturbation, or the greedy algorithm itself, any change in routing leads to the evaluation, responsible for recalculating the relevant solution information.

Upon evaluation of a call, the arrival and operation times of the vessel in the port are calculated considering berth constraints and avoiding potential infeasibility. If the operation occurs within acceptable limits and no constraint is violated, then no penalty is incurred, and inventory levels at the ports are updated accordingly. However, if a violation occurs prior to the scheduled operation, penalty costs are applied, and inventory levels are corrected to the admissible limit (upper or lower), depending on the port type (production or consumption).

Subsequently, routing costs are calculated and the planning horizon is checked. Calls exceeding the horizon are discarded, and the final inventory violations and total routing costs are recalculated. At the end of each evaluation, all relevant information is updated to ensure consistency, allowing for the next evaluation in upcoming movements.

\subsection{Beam Search Algorithm}
\label{sec: Beam Search Algorithm}

The Harpy Speech Recognition System \citep{lowerre1976harpy} was the first to implement what later became known as Beam Search \citep{OW01011988}, introducing a heuristic-based approach to efficiently explore and prune the search space in speech recognition tasks. Although the procedure was originally referred to as the Locus Model of Search, the term Beam Search was already used by \cite{Reddy}. BS is a method for exploring decision trees, especially when dealing with a large solution space. It involves methodically expanding a limited set of solutions simultaneously, aiming to increase the chance of finding a high-quality solution with limited searching effort.

The BS algorithm, as described in \cite{morais2024exact}, is derived from the concept of breadth-first search. It explores a graph by expanding the most promising nodes while maintaining a fixed number of nodes at each level. The approach is structured into six stages: Initialization, Node Expansion, Successor Evaluation, Beam Selection, Iteration, and Termination and Output generation.

During the Initialization stage, the algorithm establishes the starting parameters, including the initial node, which will always be an empty list, alongside the beam width. In the Node Expansion phase, successors are formed for each active node. Each successor is a partial solution. For a given node and its corresponding solution, all valid calls that can be appended to the solution are considered. Then, for every call, a new node is created that incorporates all previous calls together with the new one. Consequently, at the $i$-$th$ level of the generated tree, every solution consists of $i$ calls. In the Successor Evaluation phase, a greedy heuristic function is employed to assess the quality of each successor. Next, the Beam Selection phase involves selecting the most promising successors based on the heuristic evaluations while adhering to the beam width limit. The process continues with the Iteration stage, repeating until no further successors can be generated from the final level within the planning horizon. In the Termination and Output generation phase, the best solutions generated in all steps are selected.

In the present study, the BS heuristic is configured to accept three different parameters as input, which have an impact on different phases of the BS procedure. The parameters are the number of nodes allowed at a given level (\textit{N}), the number of greedy solutions for each node (\textit{q}) and the maximum number of children per node (\textit{w}). 

In the context of the proposed BS algorithm, an increase in the number of evaluated nodes, driven by a deeper or wider search tree, directly impacts CPU time. Consequently, the initial parameter to consider is $N$, which represents the number of nodes allowed within a specified level. The challenge lies in selecting a value that ensures an adequate array of potential initial solutions while keeping the tree's size within reasonable limits, which would otherwise result in higher CPU time. This effect becomes even more critical as the problem instance increases in complexity, particularly with the addition of more ports, vessels, or a longer planning horizon. Although a more extensive exploration of the solution space may be acceptable for small instances, it can quickly become computationally prohibitive for larger ones. Therefore, careful calibration of the tree structure is essential to ensure a proper balance between solution quality and computational efficiency. 

The Successor Evaluation phase follows the Node Expansion phase. Comparing partial solutions in the MIRP is a high challenge. Therefore, during the evaluation of successors, different criteria were tested. Comparison of incomplete solutions can lead to imprecise and unsatisfactory evaluations. As a result, a greedy heuristic is applied to evaluate the quality of each generated successor. Thus, the second parameter, denoted as $q$, specifies the number of greedy solutions generated during the process. Specifically, the greedy heuristic is executed once following a deterministic approach, and subsequently, it is repeated $q-1$ times incorporating randomness into its execution to introduce variability in the solution space. To select the best successors, the median of the associated $q$ values is used as a representative measure. The third parameter, denoted as $w$, represents the upper limit on the number of children that can be generated by a single node. This parameter plays an important role in maintaining the diversity and heterogeneity among the solutions. 

One problem already mentioned in Section \ref{sec:Representation} is the fact that several identical solutions are represented in different ways. A simple procedure was devised with the aim of reducing the occurrence of solutions that share the same cost, thereby improving the efficiency of BS, as it is undesirable to occupy unnecessary space in the BS level. This procedure functions by keeping a list in which the keys correspond to the scores of potential solutions. Referencing this list, it is possible to identify whether a particular score has been assigned to any solution. In such cases, these solutions are excluded from further consideration. An unlikely issue may arise with the presence of two distinct solutions that share the same cost, due to the intricate nature of the problem. Nevertheless, even if this occurs, the benefits continue to outweigh the disadvantages.

\subsection{The Greedy Algorithm}
\label{subsec: Greedy}

To effectively compare partial solutions, i.e., the successors produced during a BS execution, a greedy heuristic is used to construct a complete solution starting from a given partial solution, as presented in Algorithm \ref{alg.greedy}. The heuristic builds a feasible solution using a greedy method, making each decision by selecting the most promising immediate choice according to a defined criterion. The criterion used involves choosing the next port at risk of violating inventory levels and identifying which vessel can arrive there first. Therefore, the path is designed to prioritize, in the subsequent call, the inclusion of the port at risk of inventory violation and the vessel capable of reaching this port in the earliest possible time.

\begin{algorithm}[htbp]
\footnotesize
\renewcommand{\gets}{\leftarrow}
\SetAlgoLined
    
    $s_{partial} \gets$ BS(instance, time\_horizon, $N$, $w$)
    
    \While{ $t \leq$ time\_horizon } {
        next\_port $\gets$ \textit{next port to violate}
        
        next\_vessel $\gets$ \textit{vessel arrives first}
        
        $s_{partial} \gets s_{partial}\ \cup $ (next\_port, next\_vessel)
    }
    
    $s \gets s_{partial}$
    
    \textbf{return} $s$
            
\caption{Greedy Algorithm - Deterministic version} 
\label{alg.greedy}
\end{algorithm}

To enhance flexibility and solution exploration, a nondeterministic version of the greedy algorithm is implemented, introducing randomness at two stages of the process, the port selection and the vessel selection. This randomness mechanism allows the algorithm to explore a wider range of possible paths, helping to diversify the solutions generated by the constructive method and potentially increasing the chances of finding better solutions compared to the deterministic version. Instead of choosing the port that first violates its capacity constraint or the ship that first serves the selected port, ports with earlier violation times and vessels with earlier arrival times have a higher probability of being selected. 

By incorporating a probability factor within a normal distribution and considering the standard deviation, it is possible that the next port on the route might not be the first to face inventory violation, and the selected vessel may not be the one that reaches there first. The normal distribution was used to sample values around the average, introducing variation but ensuring that the majority of samples close to that remained near the central value.

Notably, throughout the BS procedure, multiple complete solutions are built using this greedy heuristic. To ensure that valuable solutions are kept, we keep the best $N$ solutions identified by the greedy heuristic, in case they outperform the final solution of the BS.

\subsection{Local search}
\label{sec:Neighborhoods}

The local search procedure is based on a Randomized Variable Neighborhood Descent (RVND) scheme, where each neighborhood is explored in sequence until an improvement is found. If no improvement is found, the procedure moves to the next neighborhood, following a random order. If any neighborhood results in a better solution, the process is restarted. Furthermore, our implementation uses the first improvement strategy.

An important consideration when considering these neighborhoods is that in some cases certain calls may be displaced outside the planning horizon. When this issue occurs, these calls are ignored during the solution evaluation, as they do not contribute to the cost or feasibility within the planned period.


Six neighborhood structures are implemented, as described as follows.

\begin{itemize}
    \itemsep0em
    \item \texttt{Swap}: swap two \textit{calls} in the solution. 
    \item \texttt{Relocate}: a \textit{call} is relocated to a new position in the solution. 
    \item \texttt{Replace}: the port of a \textit{call} is replaced by a port of the same type.
    \item \texttt{Insert}: a pair of \textit{calls} is inserted at the end of the solution.
    \item \texttt{Remove}: removes a \textit{call} along with the next \textit{call} of the same vessel. This procedure is carried out to preserve vessel parity. For example, if the \textit{call} (P3, V1) is removed, it becomes necessary to remove the next call of V1 to maintain the producer-consumer balance.
    \item \texttt{Swap Port}: the destinations of two vessels are swapped, while restricting the swap to ports of the same type.
    
\end{itemize}

\subsection{Iterated Local Search}
\label{sec:ILS}


ILS \citep{lourencco2003iterated} applies an iterative process that alternates between perturbation and local search procedures. In each iteration, a perturbation is randomly applied to the current solution, which involves making a controlled change in its structure, by altering the routing structure of the current solution, triggering a change in the cost value of the solution. The perturbation mechanism selects one of the six neighborhoods defined in Section \ref{sec:Neighborhoods} and applies a random move. In our implementation, such a mechanism is called twice in a row. This perturbation generates a new candidate solution which is then refined using a local search to improve its quality within the neighborhood of the current solution. The algorithm then checks if the new solution will be accepted by a pre-established Simulated Annealing criterion \citep{kirkpatrick1983optimization}. Once a solution is accepted, the algorithm updates the current solution and checks for improvement. If no improvement occurs, a counter increments. When this counter exceeds the allowed number of non-improving iterations, the solution is restored, and the counter resets. This process repeats until the maximum number of iterations is reached, at which point the final solution is returned.

%% file: sections/05.ComputationalExperiments.tex
\section{Computational Experiments}
\label{sec:results}

The proposed algorithm was coded in Julia 1.10.5 and executed on an Intel(R) Core(TM) i7-8700K CPU @ 3.70GHz running Ubuntu 20.04 on a single thread.

\subsection{Instance Details}
\label{sec:instances}

Benchmark instances and their best-known solutions are maintained on the MIRPLib website\footnote{\url{http://mirplib.scl.gatech.edu}}. The metadata includes high-level information for each instance. Most of the parameters are self-explanatory and were previously presented in Section \ref{sec:probldesc}. The reader is referred to \cite{papageorgiou2014mirplib} for further explanation. A software package has been created to read and use these public instances. Upon publication of the article, this package will be published.

This paper addresses Group 2 instances, representing a single-product MIRP, considering only one port per region and never involving split pickups or deliveries. All instances have one loading region (LR) and different numbers of discharging regions (DR). For example, instance \texttt{LR1DR02VC01V6a} has one loading region, two discharging regions, six vessels of one class, while instance \texttt{LR1DR04VC03V15b} has one loading region, four discharging regions, 15 vessels of three classes. The final letter ``a'' or ``b'' refers to the production/demand rates in the ports. If the letter is ``a'' the rates are constant, and if it is ``b'' the rates vary during the time horizon.

According to the classification presented by \cite{papageorgiou2018recent}, instances are grouped into three distinct categories based on their difficulty: Easy (E), Medium (M), and Hard (H). An instance is categorized as hard if, within a time limit of 1800 seconds, no commercial MILP solver operating with default parameter settings obtains a reduction greater than 10\% of the gap. In contrast, an instance is considered easy if at least one commercially available MILP solver can reduce the gap by more than 90\% within the same time limit using the default solver settings. Instances that do not meet the criteria for the easy or hard categories are classified as medium.




\subsection{Parameter Tuning}
\label{sec:calibration}

\subsubsection{BS Parameters}
\label{sec:BSparam}

To establish a promising value for the maximum number of children per node $w$, a set of experiments was carried out. The tests were carried out on easy instances, with $N = 100$ and 10 seeds per instance. Table \ref{tab:w} reports the average and median objective value gaps obtained from tests in which the parameter $w$ was changed from 2 to 7. The percentages represent the aggregated gap in the objective values across all runs. The best performance was achieved with $w = 2$, which resulted in the lowest average and median gaps.

\begin{table}[htbp]
\centering
\footnotesize
\caption{Gap analysis for different $w$ values.}
\label{tab:w}
\begin{tabular}{ l c c c c c c c}
\toprule
\textbf{$w$} & \textbf{2} & \textbf{3} & \textbf{4} & \textbf{5} & \textbf{6} & \textbf{7} \\ \midrule
\textbf{Average} & \underline{\text{-0.70\%}} & \text{-0.44\%} & \text{1.01\%} & \text{0.60\%} & \text{-0.21\%} & \text{-0.26\%} \\ 
\textbf{Median} & \underline{\text{-0.13\%}} & \text{-0.01\%} & \text{0.05\%} & \text{0.00\%} & \text{0.00\%} & \text{0.00\%} \\ \bottomrule
\end{tabular}
\end{table}

\subsubsection{Greedy Heuristic Parameters}
\label{sec:Greedyparam}

\begin{figure}[htbp] 
    \centering
	\includegraphics[trim=50 70 55 60,clip,width=0.7\textwidth]{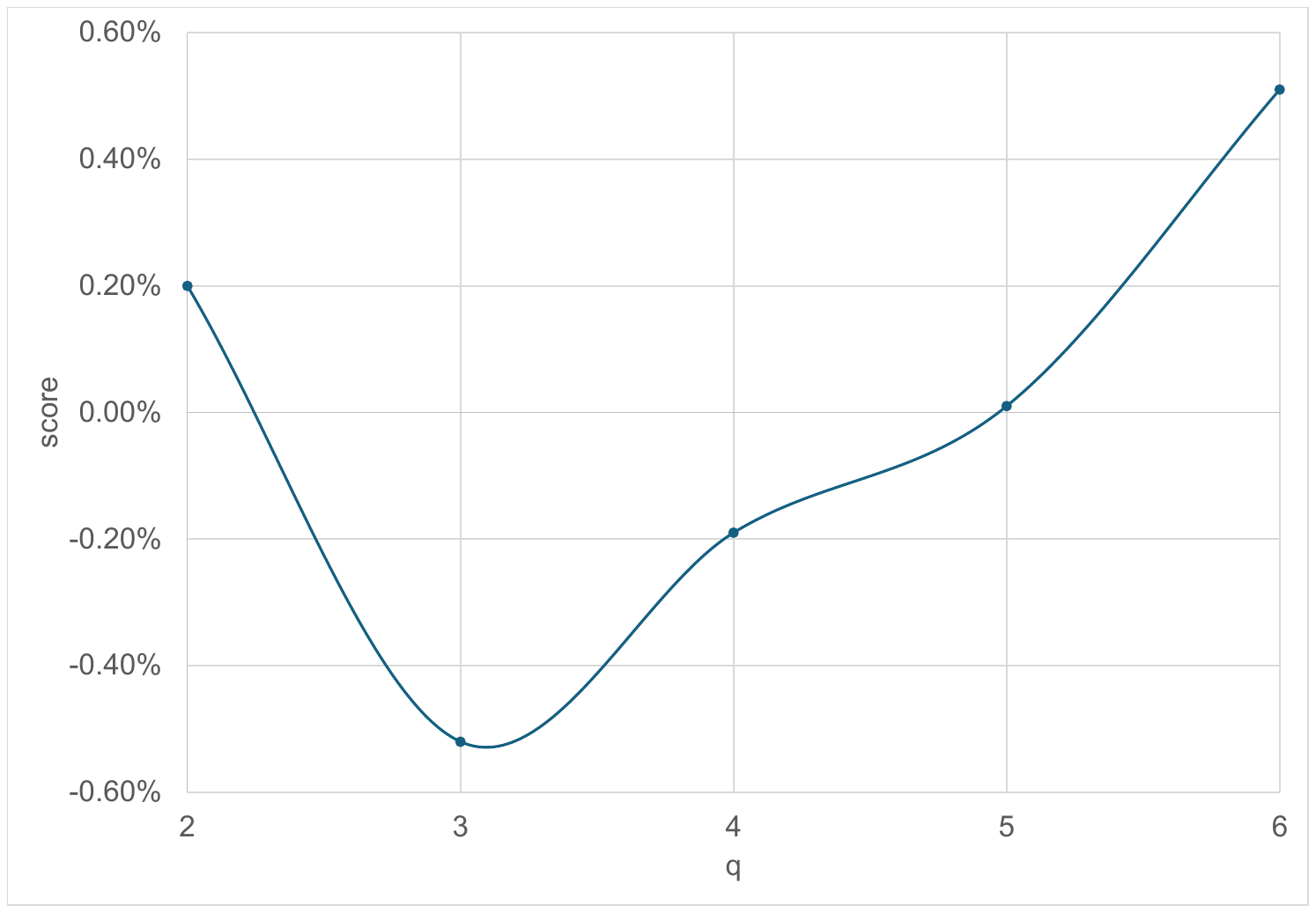}
	\caption{Visual representation of the average score across instances. \label{fig:q}}
\end{figure}

Further tests were carried out to obtain the number of greedy solutions $q$ for each node, with the aim of maximizing performance relative to CPU time. To this end, the number of nodes allowed per level in the BS $N$ was set to be inversely proportional to $q$, ensuring that the CPU time remained approximately constant. Figure \ref{fig:q} illustrates the results, showing the calibration of $q$ for easy instances. The score indicates the average gap obtained across 10 different seeds. The best results were achieved for $q = 3$, as it yielded the smallest average gap.

\subsubsection{ILS Parameters}
\label{sec:ILSparam}

To tune the ILS parameters, the Julia \texttt{HyperTuning.jl} package \citep{MejadeDios2021} was used for automated hyperparameter optimization. We selected 15 instances and five seeds, conducting a total of 120 trials that lasted 72 hours. The parameter range, step, and results are detailed in Table \ref{tab:Param opt}.  The results indicated the final set of parameter values that corresponds to the configuration that achieved the best overall performance.

\begin{table}[htbp]
    \centering
    \footnotesize
    \caption{ILS Parameter Optimization}
    \label{tab:Param opt}
    \resizebox{0.85\textwidth}{!}{

    \begin{tabular}{cccc}
    \toprule
       \textbf{Parameter} & \textbf{Range} & \textbf{Step} & \textbf{Result} \\ \midrule
        SA initial probability & 0.75--0.99 & $0.01$  & $0.79$ \\
        SA final probability & 0.01--0.10 & $0.01$ & $0.01$ \\
        ILS number of iterations & 10--1000 & 10 & 640\\
        ILS number of non-improving iterations before solution restore & 1--20 & 1 & 4\\
        ILS number of perturbations & 1--5 & 1 & 2\\
        \bottomrule
    \end{tabular}
    }
\end{table}

\subsection{Results Analysis}
\label{sec:analysis}


\subsubsection{Proposed Heuristic Results}
\label{sec:heuristicresults}

Two studies \citep{papa2013,papageorgiou2018recent} used MIRPLib Group 2 instances. In the first study, the authors first performed a warm-up phase to obtain an initial good solution, which was then used to warm-start the solver. Subsequently, the main experiments were carried out for 24 hours. In the following research \citep{papageorgiou2018recent}, the overall CPU time per scenario remains unclear, due to the combination of multiple techniques with different times. Moreover, \cite{papa2013} emphasizes that it is difficult to find good solutions for Group 2 in a few minutes, establishing a balance between the quality of the results and the time required to obtain them. Therefore, this work aims to evaluate how the heuristic works under different CPU times. To accomplish this, the parameter $N$ for the BS is evaluated using three different values ($N \in \{10, 100, 1000\}$). For each $N$, if the instance does not reach 24 hours of CPU time, it moves to the next value of $N$. If it exceeds 24 hours, it does not proceed to the next value of $N$. This approach is intended to analyze how results improve with increased CPU time using the work by \cite{papageorgiou2018recent} as a baseline reference. Another important point is that the gap reported in our results is the best gap obtained from the executed runs in 90,000 seconds. If any stage of the main algorithm was not completed within this time limit, the reported result corresponds to the last completed stage. For example, if the algorithm does not finish the ILS phase, the reported value reflects the best result achieved up to the LS, along with its respective time. Similarly, if the algorithm does not complete the LS, the reported result comes from the BS, including both the objective value and the time.

\begin{figure}[htbp]
    \centering
	\includegraphics[width=0.7\textwidth]{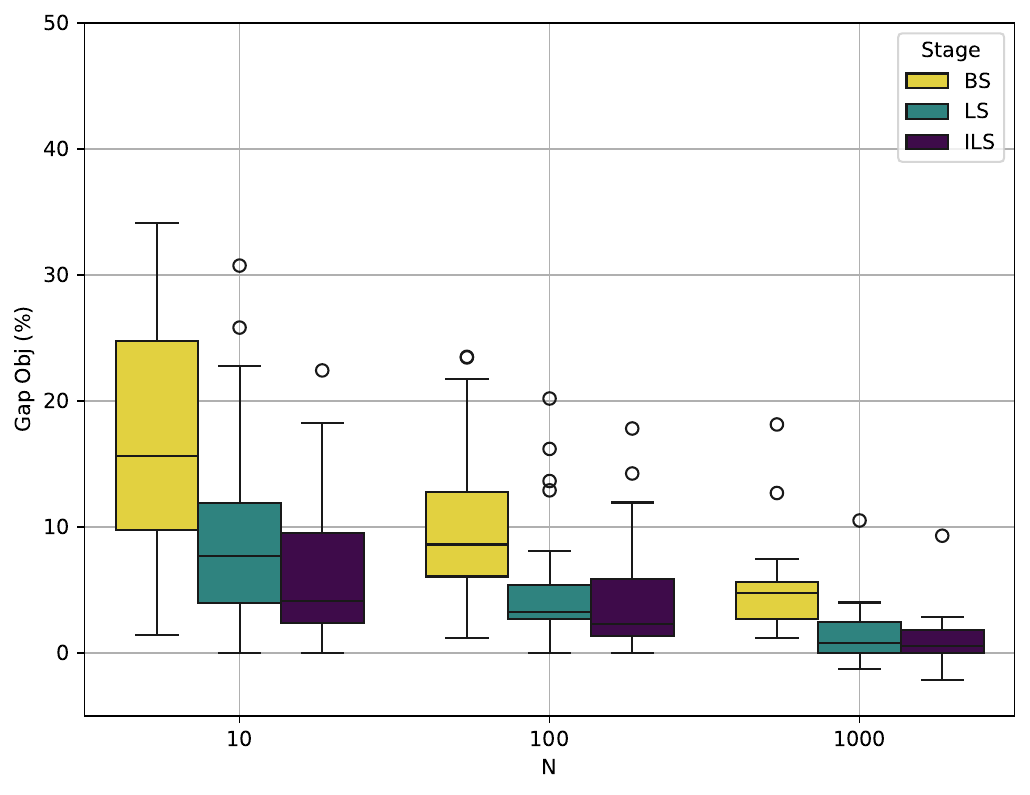}
	\caption{H = 120 with different $N$. \label{fig:120results}}
\end{figure}

\begin{figure}[htbp]
    \centering
	\includegraphics[width=0.7\textwidth]{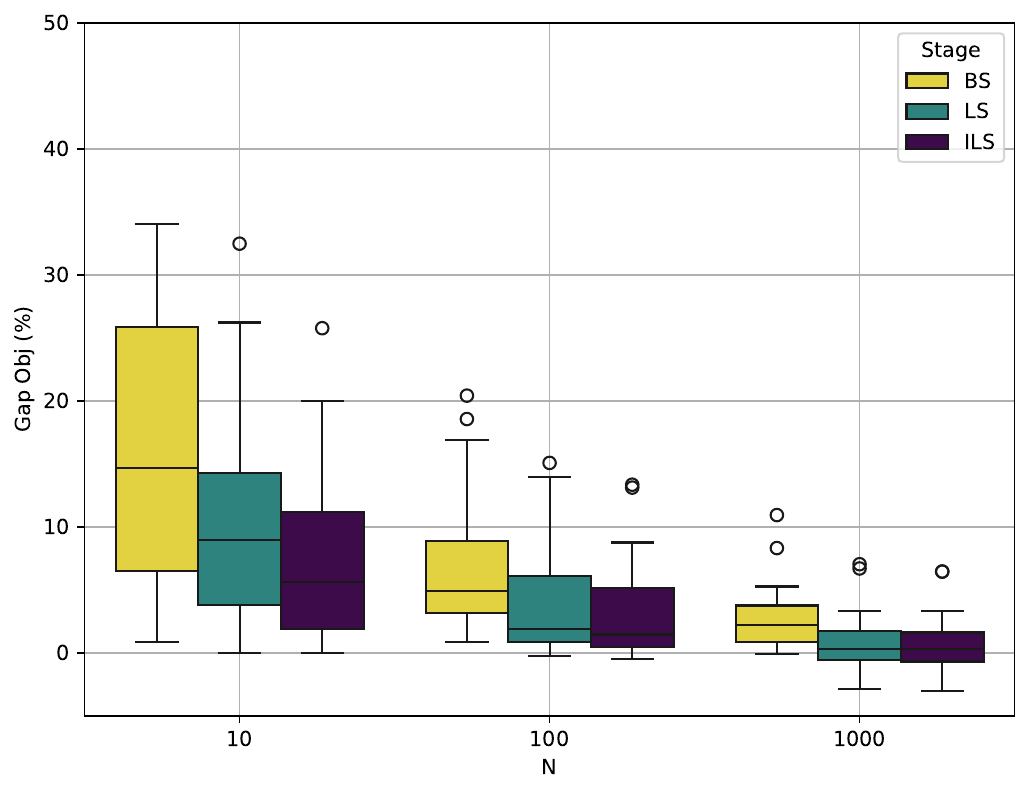}
	\caption{H = 180 with different $N$. \label{fig:180results}}
\end{figure}

\begin{figure}[htbp]
    \centering
	\includegraphics[width=0.7\textwidth]{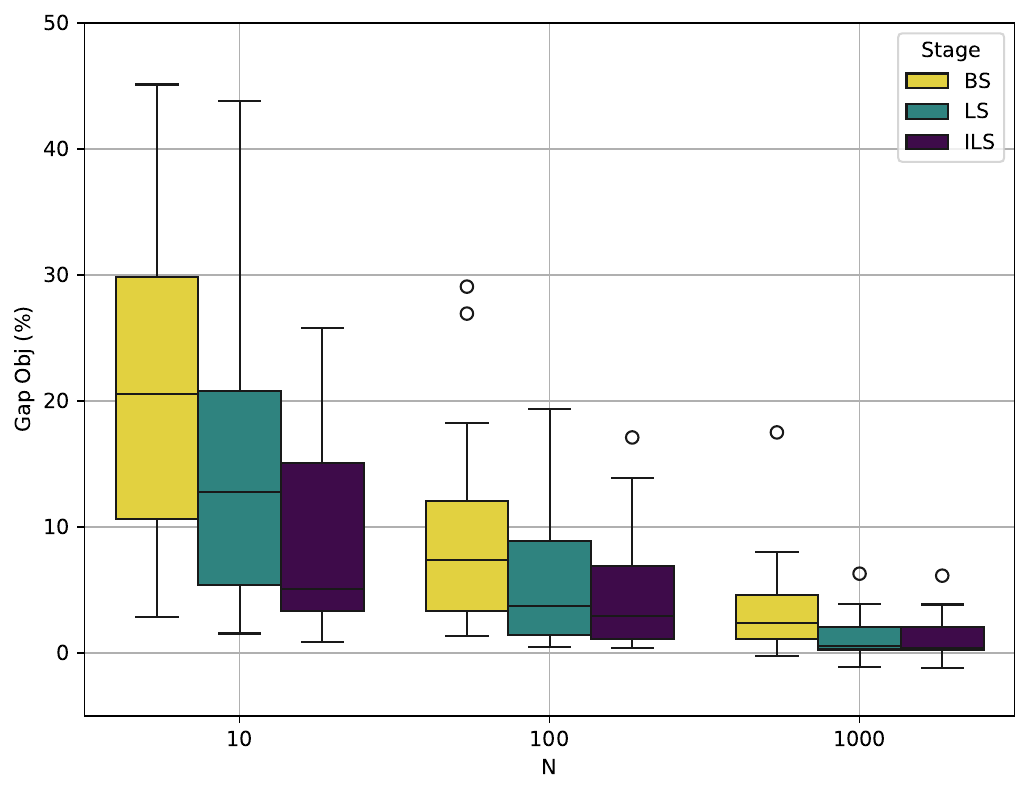}
	\caption{H = 360 with different $N$. \label{fig:360results}}
\end{figure}

The box plots presented in Figures \ref{fig:120results}, \ref{fig:180results}, and \ref{fig:360results} illustrate the comparison between different stages (BS, LS, and ILS) and the variable $N$ over three distinct values (10, 100, and 1000), providing relevant information on the influence of the parameter $N$ on the results. The y-axis represents the percentage of the objective gap (Gap Obj), while the x-axis denotes the different values of $N$.

A general observation indicates that as the value of $N$ increases from 10 to 1000, there is a visible decrease in the median Gap Obj across all stages, particularly in the BS stage, which is directly affected by this parameter. This suggests that larger values of $N$ are associated with lower percentages of gaps. Furthermore, it can be concluded that while LS has a considerable influence in all the cases analyzed, the effect of ILS tends to decrease as the value of $N$ increases. This observation remains consistent across various instance sizes, with the only notable variation being the increase in Gap Obj for larger instances. Furthermore, the interquartile ranges for each stage illustrate that there is a significant reduction in variability as $N$ increases. At $N = 10$, all stages exhibit a wide spread in values, with ILS showing the smallest variability. As $N$ increases, the number of outliers decreases, which indicates more consistent performance between stages. 

\begin{figure}[htbp]
    \centering
    \includegraphics[width=0.7\textwidth]{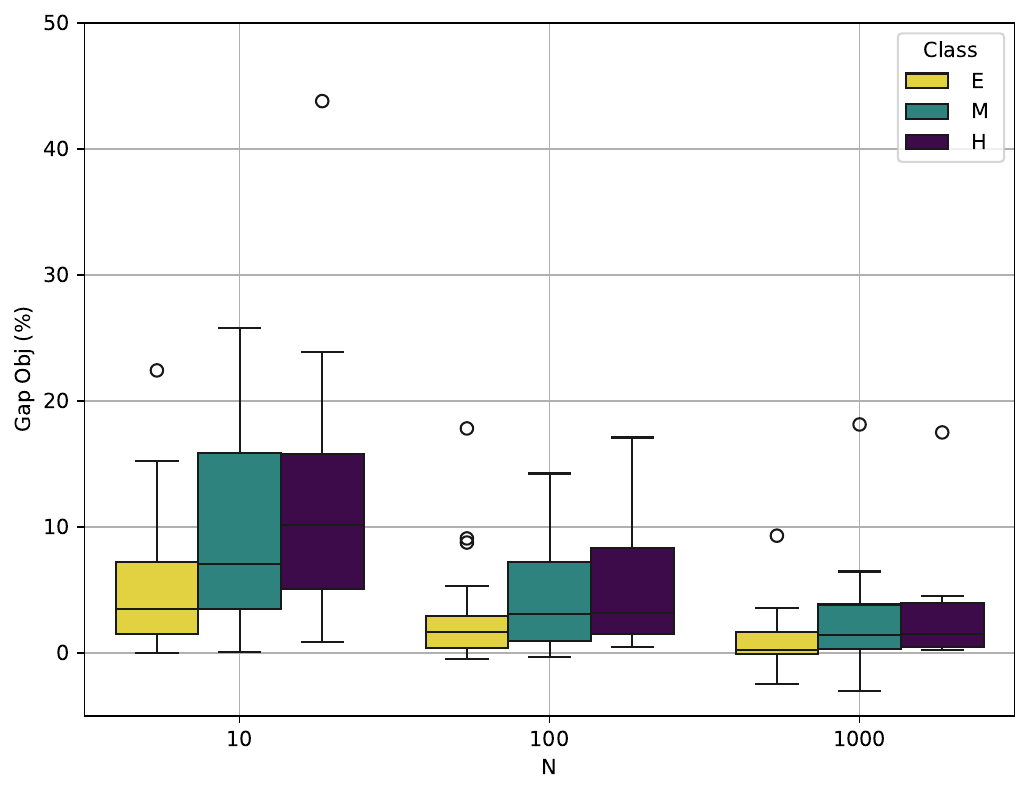}
	\caption{Gap Obj by class and $N$.\label{fig:E_M_H_results}}
\end{figure}

\begin{figure}[htbp]
    \centering
	\includegraphics[width=0.7\textwidth]{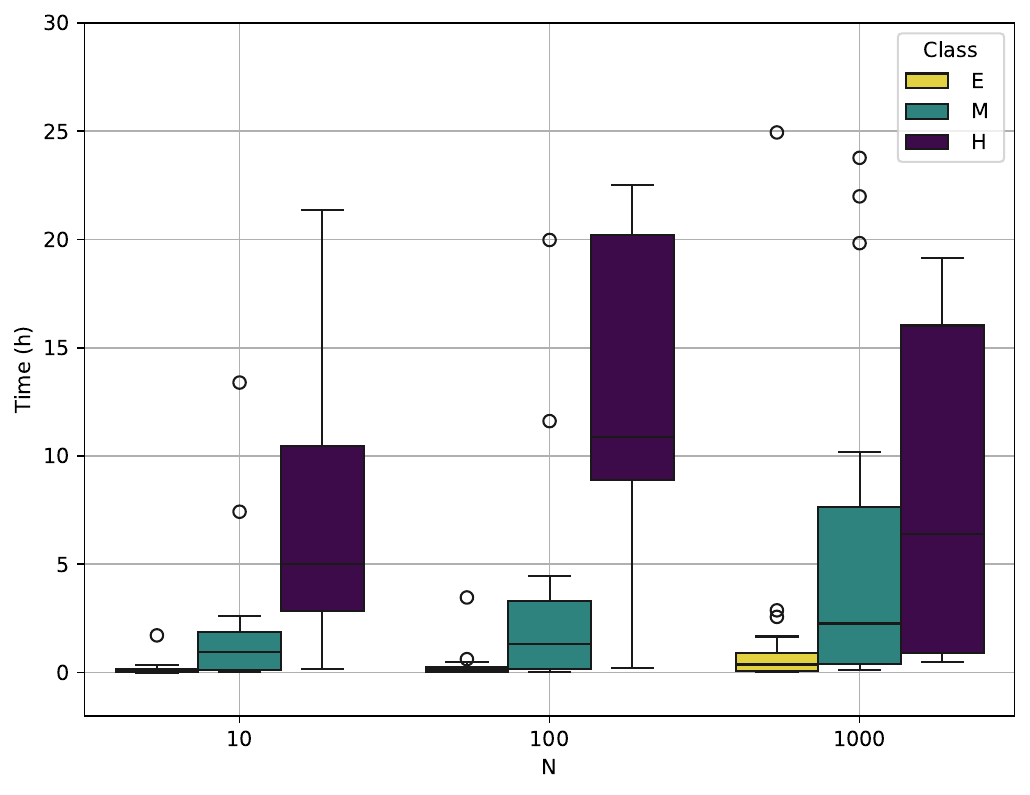}
	\caption{Time by Class and N.\label{fig:timeresults}}
\end{figure}

The box plots presented in Figures \ref{fig:E_M_H_results} and \ref{fig:timeresults} illustrate the distribution of Gap Obj and time between different values of $N$ and the classifications of instances. 
For Gap Obj, instances classified as easy consistently exhibit lower values, the median decreasing as $N$ increases. These instances are relatively easier to solve, achieving near-optimal solutions with a lower deviation. The medium class presents higher variability, especially for lower values of $N$, with a gradual reduction in dispersion as the size of the problem increases. In contrast, the hard class maintains a moderate but stable Gap Obj distribution, with values slightly decreasing as $N$ grows, although with less visible variability compared to the medium class.

With respect to time, there is a clear increase in computational effort as $N$ grows. For the easy class, CPU times remain low in all values of $N$, indicating efficient convergence. In the medium class, CPU times increase significantly, especially for higher $N$ values ($N$ = 1000), where a wider spread suggests inconsistent solver performance between different instances. The hard class consistently requires the highest computational effort, and as $N$ grows, there is a substantial increase in CPU time, confirming the difficulty of these instances. The CPU time distribution in the hard class also exhibits larger variability, suggesting that some instances require substantially more computational resources.

The main results for instances with 120, 180, and 360 periods are summarized in Tables \ref{tab:results120}, \ref{tab:results180}, and \ref{tab:results360}, respectively. The tables include key metrics such as the value of $N$, the classification of each instance (\texttt{Class}), the objective values (\texttt{Obj}), the best total cost among the ten seeds (\texttt{Best Total Cost}), the best percentage of the objective gap (\texttt{Best Gap}), the average total costs (\texttt{Average Total Cost}), the average percentage of the objective gap (\texttt{Average Gap}) and the CPU time (in hours). These tables show the best results obtained in 90,000 seconds (approximately 25 hours). The reported results represent the best values obtained across all values tested for $N$. The cases indicated by (*) show results for BS and LS, as the ILS did not complete within 90,000 seconds. The proposed heuristic demonstrated improvements in the objective value of MIRPLib \citep{papageorgiou2018recent} in 19 different instances when $N$ = 1000, considering the best seed of each instance. As expected, increasing the number of nodes improves the results. However, this improvement is accompanied by a trade-off, specifically an increase in CPU time. 
Furthermore, regarding the classification of instances, it can be observed that those classified as easy generally exhibit lower objective gaps compared to those classified as medium or hard. 
Complete results can be found in the appendix.

\begin{table}[htbp]
\centering
\caption{120-period results}
\label{tab:results120}
\resizebox{\textwidth}{!}{
\begin{tabular}{lccrrrrrr}
\toprule
\textbf{} & \textbf{} & \textbf{} & \textbf{} & \multicolumn{1}{c}{\textbf{Best}} & \multicolumn{1}{c}{\textbf{Best}} & \multicolumn{1}{c}{\textbf{Average}} & \multicolumn{1}{c}{\textbf{Average}} & \multicolumn{1}{c}{\textbf{Time}} \\ 
\multicolumn{1}{c}{\textbf{Instance}} & \multicolumn{1}{c}{\textbf{N}} & \multicolumn{1}{c}{\textbf{Class}} & \multicolumn{1}{c}{\textbf{Obj}} & \multicolumn{1}{c}{\textbf{Total Cost}} & \multicolumn{1}{c}{\textbf{Gap (\%)}} & \multicolumn{1}{c}{\textbf{Total Cost}} & \multicolumn{1}{c}{\textbf{Gap (\%)}} & \multicolumn{1}{c}{\textbf{(h)}} \\ 
\midrule
LR1\_DR02\_VC01\_V6a & 1000 & E & 33,809.00 & 33,808.95 & 0.00 & 33,808.95 & 0.00 & 0.03 \\ 
LR1\_DR02\_VC02\_V6a & 1000 & E & 74,982.00 & 74,981.65 & 0.00 & 74,986.52 & 0.01 & 0.04 \\ 
LR1\_DR02\_VC03\_V7a & 1000 & E & 40,446.00 & 40,340.01 & -0.26 & 40,403.90 & -0.10 & 0.05 \\ 
LR1\_DR02\_VC03\_V8a & 1000 & E & 43,721.00 & 43,721.43 & 0.00 & 43,933.34 & 0.48 & 0.04 \\ 
LR1\_DR02\_VC04\_V8a & 1000 & E & 41,657.00 & 41,708.69 & 0.12 & 41,847.57 & 0.46 & 0.08 \\ 
LR1\_DR02\_VC05\_V8a & 1000 & E & 36,659.00 & 36,536.64 & -0.33 & 36,616.09 & -0.12 & 0.06 \\ 
LR1\_DR03\_VC03\_V10b & 1000 & M & 92,941.00 & 92,810.27 & -0.14 & 95,667.58 & 2.85 & 0.18 \\ 
LR1\_DR03\_VC03\_V13b & 1000 & E & 124,921.00 & 125,094.57 & 0.14 & 127,203.29 & 1.79 & 0.43 \\ 
LR1\_DR03\_VC03\_V16a & 1000 & E & 82,837.00 & 86,453.57 & 4.18 & 91,335.26 & 9.30 & 0.26 \\ 
LR1\_DR04\_VC03\_V15a & 1000 & E & 73,312.00 & 74,072.61 & 1.03 & 74,714.08 & 1.88 & 0.67 \\ 
LR1\_DR04\_VC03\_V15b & 1000 & E & 117,812.00 & 115,152.95 & -2.31 & 116,451.06 & -1.17 & 0.77 \\ 
LR1\_DR04\_VC05\_V17a & 1000 & E & 72,876.00 & 73,446.65 & 0.78 & 73,864.03 & 1.34 & 0.89 \\ 
LR1\_DR04\_VC05\_V17b & 1000 & E & 105,766.00 & 99,945.85 & -5.82 & 103,534.03 & -2.16 & 1.02 \\ 
LR1\_DR05\_VC05\_V25a & 1000 & E & 105,328.00 & 105,579.09 & 0.24 & 105,995.54 & 0.63 & 2.57 \\ 
LR1\_DR05\_VC05\_V25b & 1000 & E & 137,107.00 & 135,887.81 & -0.90 & 140,504.31 & 2.42 & 2.87 \\ 
LR1\_DR08\_VC05\_V38a & 1000 & M & 166,615.00 & 169,501.81 & 1.70 & 169,501.81 & 1.70 & 19.83 \\ 
LR1\_DR08\_VC05\_V40a\* & 1000 & E & 178,593.00 & 185,214.23 & 3.57 & 185,214.23 & 3.57 & 24.94 \\ 
LR1\_DR08\_VC05\_V40b & 100 & M & 200,746.00 & 222,644.62 & 9.84 & 234,070.81 & 14.24 & 4.46 \\ 
LR1\_DR08\_VC10\_V40a & 100 & M & 185,538.00 & 188,966.37 & 1.81 & 189,991.43 & 2.34 & 3.43 \\ 
LR1\_DR08\_VC10\_V40b & 100 & M & 206,315.00 & 220,751.77 & 6.54 & 234,282.23 & 11.94 & 4.40 \\ 
LR1\_DR12\_VC05\_V70a\* & 100 & H & 278,647.00 & 287,636.40 & 3.13 & 289,736.67 & 3.83 & 18.37 \\ 
LR1\_DR12\_VC05\_V70b\* & 100 & M & 308,555.00 & 316,041.88 & 2.37 & 322,301.84 & 4.27 & 19.98 \\ 
LR1\_DR12\_VC10\_V70a & 10 & H & 283,154.00 & 289,629.21 & 2.24 & 290,673.25 & 2.59 & 12.61 \\ 
LR1\_DR12\_VC10\_V70b & 10 & H & 295,126.00 & 300,647.25 & 1.84 & 304,041.16 & 2.93 & 15.71 \\
\bottomrule
\end{tabular}
}
\end{table}

\begin{table}[htbp]
\centering
\caption{180-period results}
\resizebox{\textwidth}{!}{
\label{tab:results180}
\begin{tabular}{lccrrrrrr}
\toprule
\textbf{} & \textbf{} & \textbf{} & \textbf{} & \multicolumn{1}{c}{\textbf{Best}} & \multicolumn{1}{c}{\textbf{Best}} & \multicolumn{1}{c}{\textbf{Average}} & \multicolumn{1}{c}{\textbf{Average}} & \multicolumn{1}{c}{\textbf{Time}} \\ 
\multicolumn{1}{c}{\textbf{Instance}} & \multicolumn{1}{c}{\textbf{N}} & \multicolumn{1}{c}{\textbf{Class}} & \multicolumn{1}{c}{\textbf{Obj}} & \multicolumn{1}{c}{\textbf{Total Cost}} & \multicolumn{1}{c}{\textbf{Gap (\%)}} & \multicolumn{1}{c}{\textbf{Total Cost}} & \multicolumn{1}{c}{\textbf{Gap (\%)}} & \multicolumn{1}{c}{\textbf{(h)}} \\ 
\midrule
LR1\_DR02\_VC01\_V6a & 1000 & E & 52,167.00 & 52,167.21 & 0.00 & 52,167.21 & 0.00 & 0.07 \\ 
LR1\_DR02\_VC02\_V6a & 1000 & E & 129,372.00 & 129,372.06 & 0.00 & 129,696.36 & 0.25 & 0.09 \\ 
LR1\_DR02\_VC03\_V7a & 1000 & M & 60,547.00 & 61,302.02 & 1.23 & 61,950.19 & 2.27 & 0.11 \\ 
LR1\_DR02\_VC03\_V8a & 1000 & E & 68,153.00 & 67,480.81 & -1.00 & 70,478.74 & 3.30 & 0.09 \\ 
LR1\_DR02\_VC04\_V8a & 1000 & M & 66,017.00 & 66,073.61 & 0.09 & 66,238.44 & 0.33 & 0.17 \\ 
LR1\_DR02\_VC05\_V8a & 1000 & M & 58,619.00 & 58,087.60 & -0.91 & 58,202.65 & -0.72 & 0.14 \\ 
LR1\_DR03\_VC03\_V10b & 1000 & M & 125,638.00 & 119,714.11 & -4.95 & 122,230.31 & -2.79 & 0.38 \\ 
LR1\_DR03\_VC03\_V13b & 1000 & E & 165,764.00 & 159,667.18 & -3.82 & 161,724.70 & -2.50 & 0.90 \\ 
LR1\_DR03\_VC03\_V16a & 1000 & M & 143,178.00 & 144,635.69 & 1.01 & 153,068.94 & 6.46 & 0.62 \\ 
LR1\_DR04\_VC03\_V15a & 1000 & M & 118,621.00 & 118,360.01 & -0.22 & 119,146.56 & 0.44 & 1.50 \\ 
LR1\_DR04\_VC03\_V15b & 1000 & E & 189,989.00 & 187,064.87 & -1.56 & 188,776.86 & -0.64 & 1.66 \\ 
LR1\_DR04\_VC05\_V17a & 1000 & M & 117,710.00 & 118,507.37 & 0.67 & 118,868.81 & 0.97 & 2.10 \\ 
LR1\_DR04\_VC05\_V17b & 1000 & M & 159,168.00 & 152,523.12 & -4.36 & 154,452.82 & -3.05 & 2.27 \\ 
LR1\_DR05\_VC05\_V25a & 1000 & M & 171,620.00 & 171,546.87 & -0.04 & 172,104.96 & 0.28 & 6.04 \\ 
LR1\_DR05\_VC05\_V25b & 1000 & M & 205,368.00 & 214,609.49 & 4.31 & 219,522.01 & 6.45 & 6.57 \\ 
LR1\_DR08\_VC05\_V38a & 100 & H & 274,244.00 & 276,990.76 & 0.99 & 281,247.89 & 2.49 & 9.00 \\ 
LR1\_DR08\_VC05\_V40a & 100 & H & 296,760.00 & 302,172.23 & 1.79 & 304,072.84 & 2.40 & 9.77 \\ 
LR1\_DR08\_VC05\_V40b & 100 & M & 337,559.00 & 345,775.18 & 2.38 & 354,956.21 & 4.90 & 11.61 \\ 
LR1\_DR08\_VC10\_V40a & 100 & H & 304,261.00 & 307,761.72 & 1.14 & 309,323.52 & 1.64 & 9.41 \\ 
LR1\_DR08\_VC10\_V40b & 100 & H & 331,775.00 & 347,981.67 & 4.66 & 356,027.44 & 6.81 & 11.99 \\ 
LR1\_DR12\_VC05\_V70a\* & 10 & H & 460,566.00 & 481,166.70 & 4.28 & 506,963.62 & 9.15 & 3.40 \\ 
LR1\_DR12\_VC05\_V70b\* & 10 & H & 491,160.00 & 542,281.73 & 9.43 & 573,240.49 & 14.32 & 3.77 \\ 
LR1\_DR12\_VC10\_V70a\* & 10 & H & 466,975.00 & 480,225.26 & 2.76 & 491,159.69 & 4.92 & 3.77 \\ 
LR1\_DR12\_VC10\_V70b\* & 10 & H & 470,172.00 & 483,586.40 & 2.77 & 511,421.68 & 8.07 & 4.18 \\ 
\hline
\end{tabular}
}
\end{table}

\begin{table}[htbp]
\centering
\caption{360-period results}
\resizebox{\textwidth}{!}{
\label{tab:results360}
\begin{tabular}{lccrrrrrr}
\toprule
\textbf{} & \textbf{} & \textbf{} & \textbf{} & \multicolumn{1}{c}{\textbf{Best}} & \multicolumn{1}{c}{\textbf{Best}} & \multicolumn{1}{c}{\textbf{Average}} & \multicolumn{1}{c}{\textbf{Average}} & \multicolumn{1}{c}{\textbf{Time}} \\ 
\multicolumn{1}{c}{\textbf{Instance}} & \multicolumn{1}{c}{\textbf{N}} & \multicolumn{1}{c}{\textbf{Class}} & \multicolumn{1}{c}{\textbf{Obj}} & \multicolumn{1}{c}{\textbf{Total Cost}} & \multicolumn{1}{c}{\textbf{Gap (\%)}} & \multicolumn{1}{c}{\textbf{Total Cost}} & \multicolumn{1}{c}{\textbf{Gap (\%)}} & \multicolumn{1}{c}{\textbf{(h)}} \\ 
\midrule
LR1\_DR02\_VC01\_V6a  & 1000 & E & 108,141.00  & 108,141.00  & 0.00  & 108,141.00  & 0.00  & 0.31 \\
LR1\_DR02\_VC02\_V6a  & 1000 & E & 283,031.00  & 281,693.77  & -0.47 & 283,788.36  & 0.27  & 0.44 \\
LR1\_DR02\_VC03\_V7a  & 1000 & H & 124,282.00  & 124,772.19  & 0.39  & 126,940.04  & 2.09  & 0.50 \\
LR1\_DR02\_VC03\_V8a  & 1000 & M & 141,166.00  & 142,348.28  & 0.83  & 146,807.48  & 3.84  & 0.41 \\
LR1\_DR02\_VC04\_V8a  & 1000 & M & 138,693.00  & 138,888.24  & 0.14  & 139,237.84  & 0.39  & 0.84 \\
LR1\_DR02\_VC05\_V8a  & 1000 & H & 122,598.00  & 122,520.54  & -0.06 & 122,859.71  & 0.21 & 0.71 \\
LR1\_DR03\_VC03\_V10b & 1000 & H & 259,888.00  & 255,089.04  & -1.88 & 260,807.85  & 0.35  & 1.56 \\
LR1\_DR03\_VC03\_V13b & 1000 & N & 316,441.00  & 307,047.81  & -3.06 & 312,621.28  & -1.22 & 3.88 \\
LR1\_DR03\_VC03\_V16a & 1000 & N & 327,793.00  & 338,115.45  & 3.05  & 349,216.16  & 6.13  & 3.56 \\
LR1\_DR04\_VC03\_V15a & 1000 & N & 252,710.00  & 253,395.47  & 0.27  & 256,288.64  & 1.40  & 7.75 \\
LR1\_DR04\_VC03\_V15b & 1000 & N & 339,308.00  & 336,787.96  & -0.75 & 340,710.34  & 0,41  & 7.63 \\
LR1\_DR04\_VC05\_V17a & 1000 & H & 251,623.00  & 253,554.51  & 0.76  & 254,040.64  & 0.95  & 11.23 \\
LR1\_DR04\_VC05\_V17b & 1000 & M & 304,507.00  & 306,482.10  & 0.64  & 312,046.69  & 2.42  & 10.18 \\
LR1\_DR05\_VC05\_V25a & 100  & H & 368,628.00  & 372,314.12  & 0.99  & 373,871.33  & 1.40  & 8.84 \\
LR1\_DR05\_VC05\_V25b & 100  & H & 410,053.00  & 472,870.66  & 13.28 & 494,627.59  & 17.10 & 9.79 \\
LR1\_DR08\_VC05\_V38a & 100  & H & 596,969.00  & 683,046.79  & 12.60 & 683,046.79  & 12.60 & 22.52 \\
LR1\_DR08\_VC05\_V40a & 100  & H & 652,380.00  & 654,769.74  & 0.36 & 659,967.40  & 1.15  & 20.30 \\
LR1\_DR08\_VC05\_V40b & 100  & H & 709,713.00  & 779,769.69  & 8.98  & 804,362.42  & 11.77 & 22.41 \\
LR1\_DR08\_VC10\_V40a & 100  & H & 663,245.00  & 670,198.99  & 1.04  & 672,938.28  & 1.44  & 19.92 \\
LR1\_DR08\_VC10\_V40b & 100  & H & 724,513.00  & 786,302.74  & 7.86  & 794,420.68  & 8.80  & 21.26 \\
LR1\_DR12\_VC05\_V70a & 10   & H & 1,021,389.00 & 1,125,724.51 & 9.27  & 1,193,627.62 & 14.43 & 20.07 \\
LR1\_DR12\_VC05\_V70b & 10   & H & 1,093,013.00 & 1,378,324.17 & 20.70 & 1,435,397.61 & 23.85 & 21.35 \\
LR1\_DR12\_VC10\_V70a & 10   & H & 1,024,399.00 & 1,043,792.70 & 1.86  & 1,046,678.91 & 2.13  & 19.39 \\
LR1\_DR12\_VC10\_V70b & 10   & H & 1,001,541.00 & 1,229,992.15 & 18.57 & 1,306,455.89 & 23.34 & 16.12 \\ \hline
\end{tabular}
}
\end{table}


\subsubsection{BS and Greedy Heuristic Results}
\label{sec:greedyanalysis}


As mentioned in Section \ref{subsec: Greedy}, different combinations of nondeterministic approaches were tested for the greedy heuristic: applying it separately to port selection, vessel selection, both or none. For these tests, the 13 smallest Group 2 instances were used, with four different seeds, all with a 120-day horizon, and $q$ was set to 5. The size of the instances allowed to set a large value for $N$, 1000. The configuration that achieved the best result, with the lowest aggregate average value among instances, was to apply randomness only to port selection while keeping the selection of the vessel deterministic. The second best was to apply randomness in both decisions followed by keeping greedy heuristic deterministic. The application of randomness only on vessel selection was the last. Table \ref{tab:random} (\ref{sec:sample:appendix}) presents the complete results of these tests.

Once it was established that nondeterministic methods would be used solely for the port, additional tests were performed to evaluate three distinct techniques for aggregating the results of the greedy algorithm: minimum, mean, or median cost. The experiments were then conducted for the 13 smallest instances using three horizon sizes. For each instance, four executions with different seeds were carried out. The number of nodes allowed per level in the BS was set to 200. This decrease is attributed to the fact that instances now take into account a time horizon extending up to 360 days. The method that showed the best performance was the median with the lowest aggregate average value among the instances. The second best was the mean followed by the minimum. The complete results of these tests are shown in Table \ref{tab:aggregate} (\ref{sec:sample:appendix}).

Table \ref{tab:diff greedy} highlights the importance of the BS within the algorithm. Each column corresponds to a specific algorithm stage (After BS, After LS, and After ILS), allowing a comparison of the impact of BS on the results. The percentages indicate the improvement achieved by incorporating BS as opposed to relying solely on the greedy heuristic. Comparing the results obtained through the exclusive use of greedy heuristic versus the application of the BS algorithm, it is possible to observe that the BS approach yields an average improvement of 18.40\% in the solution following the BS stage. In particular, local search has a significant impact on results in the absence of BS, reducing the advantage of employing BS to 7.14\%. When the complete heuristic is applied, the benefit of BS decreases further to 3.57\% for this set of instances. However, its use remains beneficial, despite the increased computational effort.

Comparing partial MIRP solutions can be very challenging. Different evaluation criteria were tested, but incomplete solutions resulted in ships and ports being scheduled at different times, making comparisons difficult and evaluations inaccurate. Consequently, all attempts to evaluate partial solutions without completing the total solution were found to be unsatisfactory. Partial solutions were therefore completed using the greedy algorithm, enabling meaningful comparisons within the BS framework, as shown in Table \ref{tab:beam_search_comparison}. For a balanced comparison, both the BS and the greedy heuristic were executed for equivalent CPU time. All experiments were carried out with 15 Easy instances, $N = 10$ and $10$ seeds. This approach, although computationally expensive, since it involves running a greedy algorithm for each node in the tree, demonstrated to be much more effective than other approaches for evaluating successors.

\begin{table}[htbp]
\centering
\scriptsize
\caption{Percentage differences between the BS with the greedy heuristic against the greedy heuristic }
\label{tab:diff greedy}
\begin{tabular}{c c r r r r r}
\hline
\textbf{Instance} & \textbf{Horizon} & \textbf{After BS} & \textbf{After LS} & \textbf{After ILS} \\
\hline
LR1\_DR02\_VC01\_V6a & $180$ & $0.00$ & $0.00$ & $0.00$ \\
LR1\_DR02\_VC01\_V6a & $360$ & $-17.58$ & $0.13$ & $0.13$ \\

LR1\_DR02\_VC02\_V6a & $180$ & $-4.92$ & $-0.86$ & $-0.86$ \\
LR1\_DR02\_VC02\_V6a & $360$ & $-13.76$ & $-4.72$ & $-4.17$ \\

LR1\_DR02\_VC03\_V7a & $120$ & $-15.72$ & $-5.44$ & $-5.37$ \\

LR1\_DR02\_VC03\_V8a & $120$ & $-27.07$ & $-8.52$ & $-7.65$ \\
LR1\_DR02\_VC03\_V8a & $180$ & $-25.14$ & $-1.38$ & $1.48$ \\

LR1\_DR02\_VC04\_V8a & $120$ & $-9.63$ & $-2.73$ & $-0.54$ \\

LR1\_DR02\_VC05\_V8a & $120$ & $-1.19$ & $-1.58$ & $-1.08$ \\

LR1\_DR03\_VC03\_V13b & $120$ & $-13.27$ & $-0.95$ & $-0.86$ \\

LR1\_DR03\_VC03\_V16a & $120$ & $-66.78$ & $-53.68$ & $-21.05$ \\

LR1\_DR04\_VC03\_V15a & $120$ & $-22.78$ & $-1.41$ & $0.41$ \\

LR1\_DR04\_VC03\_V15b & $120$ & $-23.91$ & $-10.11$ & $-5.12$ \\

LR1\_DR04\_VC05\_V17a & $120$ & $-3.91$ & $-0.73$ & $-0.26$ \\

LR1\_DR04\_VC05\_V17b & $120$ & $-30.25$ & $-15.05$ & $-8.64$ \\
\hline
\textbf{Average} & & -18.40 & -7.14 & -3.57 \\
\hline
\end{tabular}
\end{table}

\begin{table}[htb!]
\centering
\scriptsize
\caption{Comparison between the BS with and without the greedy heuristic in terms of percentage gap}
\label{tab:beam_search_comparison}
\resizebox{\textwidth}{!}{
\begin{tabular}{ccrrrcrrr}
\toprule
\multirow{2}{*}{\textbf{Instance}} & \multirow{2}{*}{\textbf{Horizon}} & \multicolumn{3}{c}{\textbf{BS without greedy algorithm}} && \multicolumn{3}{c}{\textbf{BS with greedy algorithm}} \\ \cmidrule{3-5}\cmidrule{7-9}
\textbf{} & & \textbf{After BS} & \textbf{After LS} & \textbf{After ILS} && \textbf{After BS} & \textbf{After LS} & \textbf{After ILS}  \\ \midrule
    LR1\_DR02\_VC01\_V6a & 180 & 79.44 & 43.88 & 5.75 && 0.90 & 0.00 & 0.13 \\ 
    LR1\_DR02\_VC01\_V6a & 360 & 79.07 & 41.66 & 5.57 && 1.76 & 0.13 & 0.13 \\ 
    LR1\_DR02\_VC02\_V6a & 180 & 82.93 & 19.94 & 30.00 && 0.83 & 0.28 & 0.28 \\ 
    LR1\_DR02\_VC02\_V6a & 360 & 89.47 & 18.06 & 28.33 && 1.67 & 0.28 & 0.28 \\ 
    LR1\_DR02\_VC03\_V7a &  120 & 93.21 & 43.64 & 27.01 && 4.72 & 0.92 & 0.93 \\ 
    LR1\_DR02\_VC04\_V8a & 120 & 49.69 & 10.14 & 8.90 && 6.90 & 0.92 & 0.92 \\ 
    LR1\_DR02\_VC04\_V8a & 180 & 92.41 & 21.20 & 20.90 && 10.16 & 1.65 & 0.13 \\ 
    LR1\_DR02\_VC04\_V8a & 360 & 94.21 & 17.09 & 19.64 && 9.20 & 0.70 & 0.00 \\ 
    LR1\_DR02\_VC03\_V12a & 120 & 94.74 & 87.61 & 42.97 && 0.92 & 0.92 & 0.92 \\ 
    LR1\_DR03\_VC08\_V13a & 120 & 87.05 & 68.06 & 38.26 && 10.81 & 2.19 & 2.19 \\ 
    LR1\_DR04\_VC02\_V15a & 180 & 91.57 & 60.19 & 33.90 && 19.25 & 2.13 & 2.13 \\ 
    LR1\_DR04\_VC03\_V15b & 120 & 93.51 & 76.33 & 39.75 && 0.70 & 2.14 & 2.14 \\ 
    LR1\_DR04\_VC05\_V17a & 120 & 94.68 & 91.77 & 41.88 && 4.66 & 0.95 & 0.03 \\ 
    LR1\_DR04\_VC05\_V17b & 120 & 93.47 & 90.82 & 46.87 && 2.22 & 0.00 & 0.00 \\ 
    \midrule
    \textbf{Average} & & 89.84 & 54.20 & 21.53 && 5.47 & 2.70 & 2.25 \\ \bottomrule
\end{tabular}
}
\end{table}

%% file: sections/06.Conclusions.tex
\section{Conclusions}
\label{sec:conc}

The main objective of this work was to develop a heuristic approach capable of solving a MIRP without relying on mathematical optimization models and therefore (commercial) MILP solvers, while also aiming to obtain high-quality solutions for MIRPLib Group 2 instances. To this end, we propose a method that combines BS with ILS. The proposed BS strategy allowed the exploration of a restricted subset of promising solution paths, significantly reducing computational effort while directing the search toward high-quality solutions. This was possible due to the implementation of a greedy algorithm, which played a crucial role in efficiently completing partial solutions during BS. Furthermore, (iterated) local search procedures demonstrated to be essential to refine and improve the quality of solutions, producing competitive results. In particular, a comparative analysis with MIRPLib benchmarks presented 19 new best-known solutions.





The proposed methodology is flexible enough to adjust its performance by modifying the number of nodes in the BS. This allows for a controlled trade-off between smaller gaps and higher computational times, or vice versa. For smaller instances, the methodology shows strong efficiency, reaching solutions with low optimality gaps within minutes, and in some cases surpassing the quality of known benchmark results. For larger instances, maintaining a low gap requires increasing computational time (through the number of nodes in the BS). Nevertheless, satisfactory results can still be achieved in less than 24 hours.

Future work includes the development of new local search operators and acceleration mechanisms to speed up the convergence of the method. In addition, adaptive perturbation mechanisms could be designed to improve the diversification phase of the proposed algorithm, helping it escape local optima.

%% file: sections/07.Acknowledgements.tex
\section*{Acknowledgments}
This work was partially supported by CAPES (Coordenacão de Aperfeiçoamento de Pessoal de Níıvel Superior - Finance Code 001), CNPq (Conselho Nacional de Desenvolvimento Científico e Tecnológico - projects 315361/2020-4, 422470/2021-0, 406245/2021-5, 309580/2021-8 and 314420/2023-1), FAPESQ (Paraíba State Research Foundation - project 041/2023) and by FAPERJ (Fundação Carlos Chagas Filho de Amparo à Pesquisa do Estado do Rio de Janeiro - projects E-26/201.417/2022, E-26/010.002232/2019 and E-26/210.041/2023).

%% file: sections/AppendixA.tex
\section{Comparison of Results: BS versus Greedy Heuristic}
\label{sec:sample:appendix}

This appendix presents the comparative results between greedy heuristic and BS, along with the experimental trials conducted to determine the appropriate type of non-determinism and to select the best aggregation strategy for the greedy heuristic.

\begin{table}[htbp]
\centering
\caption{Greedy Algorithm without BS}
\label{tab:greedy}
\resizebox{\textwidth}{!}{
\begin{tabular}{l r r r r r r r}
\toprule
\textbf{Instance} & \textbf{Horizon} & \textbf{Bound} & \textbf{Obj} & \textbf{After BS} & \textbf{After LS} & \textbf{After ILS} & \textbf{Gap Obj} \\
\midrule
LR1\_DR02\_VC01\_V6a & 180 & 52,166.00 & 52,167.00 & 52,642.82 & 52,167.21 & 52,167.21 & 0.00\% \\
 & 360 & 107,994.00 & 108,141.00 & 129,438.62 & 108,141.00 & 108,141.00 & 0.00\% \\

LR1\_DR02\_VC02\_V6a & 180 & 128,106.00 & 129,372.00 & 136,876.18 & 130,603.20 & 130,603.20 & 0.95\% \\
 & 360 & 275,347.00 & 283,031.00 & 327,394.47 & 300,509.56 & 298,807.87 & 5.57\% \\

LR1\_DR02\_VC03\_V7a & 120 & 39,318.00 & 40,446.00 & 49,126.32 & 43,045.13 & 43,018.93 & 6.36\% \\

LR1\_DR02\_VC03\_V8a & 120 & 43,717.00 & 43,721.00 & 59,676.14 & 47,892.73 & 47,508.72 & 8.66\% \\
 & 180 & 66,989.00 & 68,153.00 & 94,931.65 & 75,464.21 & 73,314.49 & 7.57\% \\

LR1\_DR02\_VC04\_V8a & 120 & 41,277.00 & 41,657.00 & 47,211.03 & 43,183.50 & 42,206.12 & 1.32\% \\

LR1\_DR02\_VC05\_V8a & 120 & 36,088.00 & 36,659.00 & 38,024.22 & 37,255.92 & 37,008.85 & 0.95\% \\

LR1\_DR03\_VC03\_V13b & 120 & 118,706.00 & 124,921.00 & 151,117.90 & 129,835.53 & 128,395.38 & 2.78\% \\

LR1\_DR03\_VC03\_V16a & 120 & 71,635.00 & 82,837.00 & 171,093.48 & 153,981.07 & 116,603.27 & 40.76\% \\

LR1\_DR04\_VC03\_V15a & 120 & 72,108.00 & 73,312.00 & 96,851.16 & 76,859.70 & 74,896.02 & 2.16\% \\

LR1\_DR04\_VC03\_V15b & 120 & 102,148.00 & 117,812.00 & 153,494.96 & 130,322.79 & 123,931.88 & 5.19\% \\

LR1\_DR04\_VC05\_V17a & 120 & 71,572.00 & 72,876.00 & 80,193.56 & 74,921.30 & 74,119.90 & 1.71\% \\

LR1\_DR04\_VC05\_V17b & 120 & 84,970.00 & 105,766.00 & 146,881.76 & 123,865.09 & 116,463.71 & 10.11\% \\
\bottomrule
\end{tabular}
}
\end{table}

\begin{table}[htbp]
\centering
\caption{Comparison of the performance of different types of non-determinism}
\label{tab:random}
\resizebox{0.85\textwidth}{!}{
\begin{tabular}{l r r r r}
\toprule
\textbf{Instances} & \textbf{No Random} & \textbf{Random Port} & \textbf{Random Vessel} & \textbf{Both Random} \\ \midrule
LR1\_DR02\_VC01\_V6a & 33,808.95 & 33,808.95 & 33,808.95 & 33,808.95 \\ 
LR1\_DR02\_VC02\_V6a & 75,403.72 & 74,981.65 & 75,403.72 & 75,403.72 \\ 
LR1\_DR02\_VC03\_V7a & 41,274.88 & 40,454.77 & 41,142.09 & 40,919.53 \\ 
LR1\_DR02\_VC03\_V8a & 45,707.88 & 43,956.88 & 44,391.83 & 44,632.84 \\ 
LR1\_DR02\_VC04\_V8a & 41,962.49 & 41,959.43 & 41,969.56 & 41,954.04 \\ 
LR1\_DR02\_VC05\_V8a & 36,677.10 & 36,767.66 & 36,718.42 & 36,692.15 \\ 
LR1\_DR03\_VC03\_V10b & 89,398.66 & 98,455.49 & 97,672.14 & 98,555.54 \\ 
LR1\_DR03\_VC03\_V11a & 130,495.12 & 127,476.55 & 129,629.45 & 128,650.85 \\ 
LR1\_DR03\_VC03\_V11b & 105,661.44 & 96,296.38 & 105,158.58 & 100,656.24 \\ 
LR1\_DR04\_VC03\_V15a & 75,696.28 & 75,133.39 & 75,451.92 & 75,484.39 \\ 
LR1\_DR04\_VC03\_V15b & 126,659.43 & 112,634.77 & 125,544.45 & 121,872.08 \\ 
LR1\_DR04\_VC05\_V17a & 74,759.18 & 74,593.48 & 74,897.26 & 74,667.23 \\ 
LR1\_DR04\_VC05\_V17b & 113,388.55 & 107,885.74 & 112,136.89 & 111,400.20 \\ \midrule
\textbf{Average} & \textbf{76,222.59} & \textbf{74,185.01} & \textbf{76,455.79} & \textbf{75,745.98} \\
\bottomrule
\end{tabular}
}
\end{table}

\begin{table}[htbp]
\centering
\caption{Comparison of results using different aggregation methods.}
\label{tab:aggregate}
\scalebox{0.8}{
\resizebox{\textwidth}{!}{
\begin{tabular}{l r r r r}
\toprule
\textbf{Instance} & \textbf{Horizon} & \textbf{Minimum Cost} & \textbf{Mean Cost} & \textbf{Median Cost} \\ \midrule
LR1\_DR02\_VC01\_V6a & 120 & 33,923.34 & 33,939.33 & 34,266.97 \\ 
LR1\_DR02\_VC01\_V6a & 180 & 52,647.27 & 52,481.66 & 52,601.11 \\ 
LR1\_DR02\_VC01\_V6a & 360 & 109,414.31 & 110,131.37 & 109,655.42 \\ 
LR1\_DR02\_VC02\_V6a & 120 & 77,395.06 & 76,818.15 & 76,624.52 \\ 
LR1\_DR02\_VC02\_V6a & 180 & 135,993.47 & 135,549.76 & 135,545.76 \\ 
LR1\_DR02\_VC02\_V6a & 360 & 304,937.43 & 303,904.98 & 302,975.50 \\ 
LR1\_DR02\_VC03\_V7a & 120 & 42,910.78 & 42,464.15 & 42,991.15 \\ 
LR1\_DR02\_VC03\_V7a & 180 & 67,109.18 & 67,550.13 & 67,528.97 \\ 
LR1\_DR02\_VC03\_V7a & 360 & 140,255.62 & 143,311.20 & 142,994.77 \\ 
LR1\_DR02\_VC03\_V8a & 120 & 47,900.23 & 48,310.53 & 48,050.30 \\ 
LR1\_DR02\_VC03\_V8a & 180 & 77,256.03 & 78,405.74 & 77,311.96 \\ 
LR1\_DR02\_VC03\_V8a & 360 & 174,802.30 & 175,016.07 & 173,030.24 \\ 
LR1\_DR02\_VC04\_V8a & 120 & 42,091.77 & 42,164.18 & 42,218.00 \\ 
LR1\_DR02\_VC04\_V8a & 180 & 68,388.41 & 67,871.17 & 67,815.13 \\ 
LR1\_DR02\_VC04\_V8a & 360 & 145,597.41 & 145,606.29 & 145,889.97 \\ 
LR1\_DR02\_VC05\_V8a & 120 & 37,002.98 & 36,954.36 & 36,988.57 \\ 
LR1\_DR02\_VC05\_V8a & 180 & 58,923.26 & 58,750.84 & 58,762.10 \\ 
LR1\_DR02\_VC05\_V8a & 360 & 124,037.47 & 123,887.84 & 123,824.35 \\ 
LR1\_DR03\_VC03\_V10b & 120 & 110,858.34 & 111,754.60 & 114,018.19 \\ 
LR1\_DR03\_VC03\_V10b & 180 & 145,883.67 & 145,080.40 & 146,118.27 \\ 
LR1\_DR03\_VC03\_V10b & 360 & 37,0964.74 & 362,337.30 & 358,205.90 \\ 
LR1\_DR03\_VC03\_V13b & 120 & 136,033.99 & 134,909.53 & 135,552.79 \\ 
LR1\_DR03\_VC03\_V13b & 180 & 173,509.93 & 172,567.81 & 172,606.80 \\ 
LR1\_DR03\_VC03\_V13b & 360 & 341,677.79 & 341,456.29 & 341,661.21 \\ 
LR1\_DR03\_VC03\_V16a & 120 & 138,716.00 & 137,684.78 & 131,597.83 \\ 
LR1\_DR03\_VC03\_V16a & 180 & 276,705.77 & 276,930.69 & 274,921.64 \\ 
LR1\_DR03\_VC03\_V16a & 360 & 682,519.75 & 682,040.67 & 682,703.66 \\ 
LR1\_DR04\_VC03\_V15a & 120 & 82,961.41 & 85,314.85 & 84,632.05 \\ 
LR1\_DR04\_VC03\_V15a & 180 & 148,329.62 & 148,596.19 & 148,713.50 \\ 
LR1\_DR04\_VC03\_V15a & 360 & 301,765.71 & 301,804.37 & 301,325.94 \\ 
LR1\_DR04\_VC03\_V15b & 120 & 143,300.31 & 143,387.59 & 143,254.13 \\ 
LR1\_DR04\_VC03\_V15b & 180 & 212,916.42 & 212,891.02 & 212,375.42 \\ 
LR1\_DR04\_VC03\_V15b & 360 & 387,058.14 & 382,095.79 & 381,325.83 \\ 
LR1\_DR04\_VC05\_V17a & 120 & 75,673.42 & 75,783.27 & 75,735.27 \\ 
LR1\_DR04\_VC05\_V17a & 180 & 122,075.49 & 121,668.81 & 122,024.03 \\ 
LR1\_DR04\_VC05\_V17a & 360 & 261,484.75 & 260,646.48 & 260,256.08 \\ 
LR1\_DR04\_VC05\_V17b & 120 & 134,376.88 & 133,850.50 & 132,773.17 \\ 
LR1\_DR04\_VC05\_V17b & 180 & 197,886.09 & 198,980.83 & 199,055.59 \\ 
LR1\_DR04\_VC05\_V17b & 360 & 412,700.92 & 408,949.66 & 409,822.29 \\ \midrule
\textbf{Average} &  & \textbf{169,179.11} & \textbf{168,765,36} & \textbf{168,403.956} \\\bottomrule
\end{tabular}
}}
\end{table}

%% file: sections/AppendixB.tex
\section{Complete Final Results}
\label{sec:sample:appendixresults}

This appendix provides the complete results from the computational experiments conducted with the proposed heuristic. Detailed results of the average total cost and time are shown for each instance, in three different time horizons for different values of $N$.

\begin{footnotesize}
\setlength{\tabcolsep}{5pt}
\begin{longtable}[htbp]{lrrrrrrrr}
\toprule
& \multicolumn{2}{c}{\textbf{After BS}} && \multicolumn{2}{c}{\textbf{After LS}} && \multicolumn{2}{c}{\textbf{After ILS}}  \\ \cmidrule{2-3}\cmidrule{5-6}\cmidrule{8-9}
\multicolumn{1}{c}{\textbf{Instance}} & \multicolumn{1}{c}{\textbf{Cost}} & \multicolumn{1}{c}{\textbf{Time(s)}} && \multicolumn{1}{c}{\textbf{Cost}} & \multicolumn{1}{c}{\textbf{Time(s)}} && \multicolumn{1}{c}{\textbf{Cost}} & \multicolumn{1}{c}{\textbf{Time(s)}} \\ \midrule
\endhead
\bottomrule \caption{Results for instance with horizon = 120 (\textit{cont...})} \\
\endfoot
\bottomrule \caption{Results for instance with horizon = 120} \label{tab:resultados120}\\
\endlastfoot
LR1\_DR02\_VC01\_V6a & & & & & & & & \\
10 & 34,284.56 & 0.61 & & 33,808.96 & 1.57 & & 33,808.95 & 16.06\\
100 & 34,284.55 & 8.13 & & 33,808.95 & 3.35 & & 33,808.95 & 16.98\\
1000 & 34,284.55 & 70.62 & & 33,808.95 & 23.18 & & 33,808.95 & 17.87\\
LR1\_DR02\_VC02\_V6a & & & & & & & & \\
10 & 78,048.16 & 0.77 & & 76,887.22 & 1.73 & & 76,869.91 & 23.41\\
100 & 75,903.99 & 11.38 & & 74,981.65 & 3.75 & & 74,981.65 & 24.69\\
1000 & 75,896.88 & 93.04 & & 74,986.52 & 24.38 & & 74,986.52 & 24.74\\
LR1\_DR02\_VC03\_V7a & & & & & & & & \\
10 & 47,374.37 & 0.89 & & 45,894.60 & 1.66 & & 45,155.71 & 25.28\\
100 & 43,143.76 & 14.32 & & 42,038.41 & 4.32 & & 41,656.67 & 26.69\\
1000 & 41,229.71 & 113.97 & & 40,408.26 & 29.97 & & 40,403.90 & 27.91\\
LR1\_DR02\_VC03\_V8a & & & & & & & & \\
10 & 52,117.13 & 0.83 & & 49,413.12 & 1.63 & & 48,163.06 & 19.74\\
100 & 48,431.66 & 12.74 & & 46,704.93 & 3.86 & & 46,169.69 & 21.30\\
1000 & 45,022.11 & 100.18 & & 43,933.34 & 26.53 & & 43,933.34 & 22.40\\
LR1\_DR02\_VC04\_V8a & & & & & & & & \\
10 & 44,951.88 & 1.36 & & 42,621.17 & 1.89 & & 42,238.81 & 42.21\\
100 & 43,428.05 & 23.38 & & 42,121.21 & 6.5 & & 41,911.72 & 45.47\\
1000 & 42,952.19 & 177.68 & & 41,906.16 & 51.21 & & 41,847.57 & 45.45\\
LR1\_DR02\_VC05\_V8a & & & & & & & & \\
10 & 38,004.70 & 1.14 & & 37,204.26 & 1.73 & & 37,007.02 & 35.53\\
100 & 37,612.54 & 19.02 & & 36,754.74 & 5.49 & & 36,718.29 & 39.49\\
1000 & 37,573.26 & 148.01 & & 36,622.13 & 40.21 & & 36,616.09 & 39.54\\
LR1\_DR03\_VC03\_V10b & & & & & & & & \\
10 & 122,609.39 & 3.46 & & 115,538.44 & 2.8 & & 113,295.03 & 108.04\\
100 & 105,428.20 & 52.69 & & 101,145.26 & 15 & & 100,567.81 & 110.79\\
1000 & 98,539.75 & 418.46 & & 95,964.03 & 129.69 & & 95,667.58 & 111.58\\
LR1\_DR03\_VC03\_V13b & & & & & & & & \\
10 & 139,702.97 & 11.65 & & 134,119.65 & 4.98 & & 129,738.55 & 224.71\\
100 & 133,767.57 & 119.1 & & 129,496.89 & 36.25 & & 128,214.31 & 233.34\\
1000 & 131,139.72 & 1,048.76 & & 127,708.77 & 279.39 & & 127,203.29 & 230.36\\
LR1\_DR03\_VC03\_V16a & & & & & & & & \\
10 & 125,766.78 & 7.81 & & 119,597.46 & 2.92 & & 106,775.58 & 143.62\\
100 & 108,280.56 & 78.3 & & 103,798.86 & 16.61 & & 100,793.05 & 143.51\\
1000 & 94,871.47 & 659.31 & & 92,562.11 & 130.7 & & 91,335.26 & 141.76\\
LR1\_DR04\_VC03\_V15a & & & & & & & & \\
10 & 84,461.46 & 19.72 & & 77,781.46 & 6.36 & & 75,525.02 & 296.91\\
100 & 79,069.31 & 172.15 & & 75,794.80 & 47.38 & & 75,019.96 & 302.16\\
1000 & 77,634.98 & 1,640.48 & & 74,919.84 & 461.55 & & 74,714.08 & 296.73\\
LR1\_DR04\_VC03\_V15b & & & & & & & & \\
10 & 140,139.76 & 24.13 & & 131,008.91 & 7.53 & & 126,556.14 & 388.82\\
100 & 127,266.91 & 206.51 & & 121,240.65 & 50.38 & & 119,902.43 & 381.60\\
1000 & 121,172.64 & 1,942.67 & & 116,840.12 & 465.61 & & 116,451.06 & 380.13\\
LR1\_DR04\_VC05\_V17a & & & & & & & & \\
10 & 78,327.31 & 26.56 & & 75,297.60 & 8.04 & & 74,284.04 & 401.14\\
100 & 77,117.73 & 236.15 & & 74,464.65 & 68.62 & & 74,018.68 & 400.30\\
1000 & 76,676.89 & 2,133.04 & & 74,122.89 & 677.08 & & 73,864.03 & 389.46\\
LR1\_DR04\_VC05\_V17b & & & & & & & & \\
10 & 125,461.17 & 34.15 & & 119,074.20 & 9.18 & & 114,069.25 & 545.55\\
100 & 115,286.57 & 267.2 & & 109,308.17 & 77.12 & & 108,207.31 & 549.08\\
1000 & 108,664.72 & 2,534.21 & & 104,446.49 & 610.35 & & 103,534.03 & 533.56\\
LR1\_DR05\_VC05\_V25a & & & & & & & & \\
10 & 111,685.70 & 82.71 & & 107,879.79 & 11.16 & & 106,129.76 & 904.30\\
100 & 109,926.92 & 725.57 & & 106,844.71 & 110.32 & & 106,208.20 & 941.64\\
1000 & 108,981.57 & 7,231.93 & & 106,163.11 & 1,102.85 & & 105,995.54 & 926.51\\
LR1\_DR05\_VC05\_V25b & & & & & & & & \\
10 & 196,054.84 & 88.05 & & 184,818.63 & 18.56 & & 161,750.38 & 1,208.53\\
100 & 168,185.06 & 826.98 & & 157,415.61 & 162.53 & & 150,807.98 & 1,243.89\\
1000 & 148,161.95 & 7,797.47 & & 142,819.71 & 1,367.55 & & 140,504.31 & 1,184.17\\
LR1\_DR08\_VC05\_V38a & & & & & & & & \\
10 & 197,239.39 & 505.41 & & 181,760.20 & 101.33 & & 174,617.70 & 5,606.91\\
100 & 177,715.18 & 5,068.98 & & 171,626.89 & 1,112.23 & & 170,153.78 & 5,755.66\\
1000 & 174,950.49 & 55,892.03 & & 170,770.36 & 10,101.59 & & 169,501.81 & 5,393.35\\
LR1\_DR08\_VC05\_V40a & & & & & & & & \\
10 & 208,065.62 & 554.35 & & 192,215.93 & 116.63 & & 185,280.57 & 5,518.14\\
100 & 196,680.93 & 5,568.35 & & 188,108.08 & 1,168.29 & & 185,935.56 & 5,753.25\\
1000 & 192,555.82 & 79,114.92 & & 185,214.23 & 10,675.62 & & -- & --\\
LR1\_DR08\_VC05\_V40b & & & & & & & & \\
10 & 293,293.36 & 674.77 & & 258,690.07 & 182.79 & & 245,531.78 & 7,189.08\\
100 & 262,252.82 & 6,914.43 & & 239,513.59 & 1,656.55 & & 234,070.81 & 7,473.46\\
1000 & 245,203.61 & 85,573.09 & & -- & -- & & -- & --\\
LR1\_DR08\_VC10\_V40a & & & & & & & & \\
10 & 212,728.01 & 561.46 & & 193,728.08 & 125.79 & & 189,663.97 & 5,573.97\\
100 & 203,767.81 & 5,456.39 & & 191,443.57 & 1,289.16 & & 189,991.43 & 5,601.75\\
1000 & 196,282.68 & 79,164.35 & & -- & -- & & -- & --\\
LR1\_DR08\_VC10\_V40b & & & & & & & & \\
10 & 308,628.52 & 648.81 & & 267,068.36 & 177.62 & & 247,671.98 & 7,408.65\\
100 & 263,649.25 & 6,566.30 & & 238,891.00 & 1,600.70 & & 234,282.23 & 7,671.71\\
LR1\_DR12\_VC05\_V70a & & & & & & & & \\
10 & 366,907.97 & 4,778.66 & & 316,282.10 & 993.93 & & 299,464.95 & 35,033.96\\
100 & 317,174.34 & 56,331.11 & & 289,736.67 & 9,784.83 & & -- & --\\
LR1\_DR12\_VC05\_V70b & & & & & & & & \\
10 & 420,658.15 & 5,162.71 & & 347,804.46 & 1,097.17 & & 337,312.30 & 41,955.17\\
100 & 356,860.49 & 61,661.63 & & 322,301.84 & 10,251.69 & & -- & --\\
LR1\_DR12\_VC10\_V70a & & & & & & & & \\
10 & 340,901.25 & 5,319.50 & & 295,743.85 & 1,254.64 & & 290,673.25 & 38,812.28\\
100 & 323,554.00 & 58,142.12 & & 292,180.46 & 12,331.28 & & -- & --\\
LR1\_DR12\_VC10\_V70b & & & & & & & & \\
10 & 400,556.07 & 5,433.52 & & 315,584.26 & 1,427.82 & & 304,041.16 & 49,687.21\\
100 & 354,136.74 & 63,738.30 & & 304,932.19 & 13,348.48 & & -- & --\\
\end{longtable}
\end{footnotesize}

\begin{footnotesize}
\setlength{\tabcolsep}{5pt}
\begin{longtable}[htbp]{lrrrrrrrr}
\toprule
& \multicolumn{2}{c}{\textbf{After BS}} && \multicolumn{2}{c}{\textbf{After LS}} && \multicolumn{2}{c}{\textbf{After ILS}}  \\ \cmidrule{2-3}\cmidrule{5-6}\cmidrule{8-9}
\multicolumn{1}{c}{\textbf{Instance}} & \multicolumn{1}{c}{\textbf{Cost}} & \multicolumn{1}{c}{\textbf{Time(s)}} && \multicolumn{1}{c}{\textbf{Cost}} & \multicolumn{1}{c}{\textbf{Time(s)}} && \multicolumn{1}{c}{\textbf{Cost}} & \multicolumn{1}{c}{\textbf{Time(s)}} \\ \midrule
\endhead
\bottomrule \caption{Results for instance with horizon = 180 (\textit{cont...})} \\
\endfoot
\bottomrule \caption{Results for instance with horizon = 180} \label{tab:resultados180}\\
\endlastfoot
LR1\_DR02\_VC01\_V6a & & & & & & & & \\ 
10 & 52,642.82 & 1.12 & & 52,167.21 & 1.78 & & 52,167.21 & 47.80 \\ 
100 & 52,642.81 & 17.98 & & 52,167.21 & 5.69 & & 52,167.21 & 48.43 \\ 
1000 & 52,642.80 & 142.47 & & 52,167.21 & 44.72 & & 52,167.21 & 49.39 \\ 
LR1\_DR02\_VC02\_V6a & & & & & & & & \\ 
10 & 136,275.02 & 1.51 & & 134,692.78 & 1.97 & & 133,948.87 & 72.30 \\ 
100 & 131,981.58 & 24.47 & & 131,060.33 & 7.04 & & 131,044.55 & 74.20 \\ 
1000 & 130,773.00 & 197.94 & & 129,696.36 & 59.04 & & 129,696.36 & 75.73 \\ 
LR1\_DR02\_VC03\_V7a & & & & & & & & \\ 
10 & 70,994.35 & 1.82 & & 68,765.03 & 2.10 & & 68,450.86 & 80.70 \\ 
100 & 66,001.10 & 29.79 & & 64,467.39 & 8.69 & & 64,430.30 & 81.53 \\ 
1000 & 63.256.10 & 236.14 & & 61,983.46 & 69.20 & & 61,950.19 & 77.59 \\ 
LR1\_DR02\_VC03\_V8a & & & & & & & & \\ 
10 & 84,017.67 & 1.71 & & 80,349.55 & 2.07 & & 79,625.62 & 63.25 \\ 
100 & 76,436.39 & 26.31 & & 74,946.37 & 7.97 & & 74,695.87 & 65.17 \\ 
1000 & 71,943.71 & 208.16 & & 70,523.00 & 60.52 & & 70,478.74 & 65.12 \\ 
LR1\_DR02\_VC04\_V8a & & & & & & & & \\ 
10 & 69,941.11 & 2.80 & & 68,251.61 & 2.50 & & 67,924.21 & 130.90 \\ 
100 & 67,689.80 & 45.00 & & 66,600.69 & 14.50 & & 66,409.01 & 129.43 \\ 
1000 & 67,471.99 & 367.49 & & 66,245.28 & 131.41 & & 66,238.44 & 127.61 \\ 
LR1\_DR02\_VC05\_V8a & & & & && & & \\ 
10 & 59,559.55 & 2.41 & & 58,792.60 & 2.28 & & 58,662.78 & 110.56 \\ 
100 & 59,250.49 & 38.72 & & 58,498.32 & 11.25 & & 58,419.33 & 118.47 \\ 
1000 & 59,102.81 & 305.93 & & 58,202.72 & 95.85 & & 58,202.65 & 116.06 \\ 
LR1\_DR03\_VC03\_V10b & & & && & & & \\ 
10 & 146,394.07 & 6.63 & & 140,138.60 & 5.29 & & 137,143.80 & 290.82 \\ 
100 & 133,642.78 & 93.25 & & 129,788.09 & 35.10 & & 129,405.47 & 300.29 \\ 
1000 & 125,562.19 & 764.48 & & 122,268.37 & 319.07 & & 122,230.31 & 281.78 \\ 
LR1\_DR03\_VC03\_V13b & & & & && & & \\ 
10 & 177,821.66 & 22.76 & & 170,402.34 & 12.50 & & 166,383.84 & 628.83 \\ 
100 & 170,358.65 & 205.85 & & 166,370.12 & 88.70 & & 164.996.76 & 629.25 \\ 
1000 & 165,891.32 & 1,846.87 & & 162,019.22 & 769.19 & & 161,724.70 & 611.59 \\ 
LR1\_DR03\_VC03\_V16a & & & & & & & & \\ 
10 & 217,091.17 & 17.61 & & 212,067.87 & 6.45 & & 192,871.76 & 460.72 \\ 
100 & 172,232.63 & 156.83 & & 166,344.51 & 54.19 & & 164,804.10 & 455.89 \\ 
1000 & 156,169.37 & 1,386.48 & & 153,462.91 & 397.12 & & 153,068.94 & 459.00 \\ 
LR1\_DR04\_VC03\_V15a & & & & & & & & \\ 
10 & 137,617.16 & 40.00 & & 129,925.54 & 15.18 & & 125,869.05 & 889.36 \\ 
100 & 124,244.15 & 338.43 & & 120,451.43 & 125.97 & & 119,551.59 & 914.99 \\ 
1000 & 122,221.97 & 3,225.96 & & 119,443.85 & 1,274.73 & & 119,146.56 & 902.38 \\ 
LR1\_DR04\_VC03\_V15b & & & & & & & & \\ 
10 & 209,509.08 & 44.91 & & 198,141.58 & 20.14 & & 193,887.99 & 1,075.15 \\ 
100 & 198,572.29 & 380.51 & & 191,821.36 & 157.32 & & 190,933.64 & 1,074.36 \\ 
1000 & 194,218.77 & 3,679.88 & & 189,204.30 & 1,277.53 & & 188,776.86 & 1,035.38 \\ 
LR1\_DR04\_VC05\_V17a & & & & & & & & \\ 
10 & 124,094.13 & 53.77 & & 120,584.85 & 20.10 & & 119,486.38 & 1,295.16 \\ 
100 & 122,440.16 & 451.34 & & 119,622.72 & 183.41 & & 119,131.82 & 1,300.36 \\ 
1000 & 121,651.03 & 4,407.27 & & 119,129.10 & 1,925.75 & & 118,868.81 & 1,233.94 \\ 
LR1\_DR04\_VC05\_V17b & & & & & & & & \\ 
10 & 181,621.65 & 57.91& & 169,971.74 & 24.56 & & 167,248.46 & 1,630.00 \\ 
100 & 166,398.15 & 490.78& & 159,955.15 & 204.63 & & 158,660.29 & 1,555.65 \\ 
1000 & 159,378.47 & 4,811.62 & & 154,757.12 & 1,790.23 & & 154.452.82 & 1,566.87 \\ 
LR1\_DR05\_VC05\_V25a & & & & & && & \\ 
10 & 180,832.49 & 150.76 & & 175,354.46 & 25.40 & & 172,672.91 & 3,073.58 \\ 
100 & 177,585.91 & 1,312.77 & & 173,178.18 & 300.10 & & 172,353.76 & 3,135.68 \\ 
1000 & 175,501.16 & 15,579.02 & & 172,185.21 & 3.029.10 & & 172,104.96 & 3,153.64 \\ 
LR1\_DR05\_VC05\_V25b & & & & & & & & \\ 
10 & 310,475.54 & 163.10 & & 278,354.05 & 72.58 & & 256,690.03 & 3,904.57 \\ 
100 & 258,067.92 & 1,485.22 & & 241,850.57 & 675.21 & & 237,004.99 & 3,911.10 \\ 
1000 & 230,590.32 & 14,580.05 & & 220,912.35 & 5,276.42 & & 219,522.01 & 3,783.77 \\ 
LR1\_DR08\_VC05\_V38a & & & & & & & & \\ 
10 & 374,986.84 & 965.00 & & 322,878.03 & 413.42 & & 308,794.22 & 19,560.84 \\ 
100 & 293,240.81 & 9,720.98 & & 284,625.41 & 3,247.97 & & 281,247.89 & 19,431.67 \\ 
LR1\_DR08\_VC05\_V40a & & & & && & & \\ 
10 & 351,294.39 & 1,018.41 & & 328,664.56 & 358.74 & & 314,715.11 & 20,240.06 \\ 
100 & 322,536.79 & 10,382.88 & & 307,338.73 & 3,608.44 & & 304,072.84 & 21,178.88 \\ 
LR1\_DR08\_VC05\_V40b & & & && & & & \\ 
10 & 453,192.08 & 1,075.42 & & 393,778.01 & 618.75 & & 377,404.28 & 25,046.17 \\ 
100 & 400,234.98 & 11,621.11 & & 360,090.24 & 5,845.84 & & 354,956.21 & 24,338.27 \\ 
LR1\_DR08\_VC10\_V40a & & & && & & & \\ 
10 & 366,614.62 & 1,013.78 & & 341,535.32 & 393.07 & & 321,955.16 & 19,970.46 \\ 
100 & 321,177.49 & 9,996.80 & & 311,088.00 & 3,848.47 & & 309,323.52 & 20,026.89 \\ 
LR1\_DR08\_VC10\_V40b & & & & && & & \\ 
10 & 456,494.15 & 1,057.38 & & 392,869.06 & 631.67 & & 373,795.14 & 25,637.22 \\ 
100 & 407,400.42 & 11,644.42 & & 366,700.05 & 5,885.47 & & 356,027.44 & 25,647.93 \\ 
LR1\_DR12\_VC05\_V70a & & & & && & & \\ 
10 & 581,917.48 & 9,023.62 & & 506,963.62 & 3,224.21 && -- & -- \\ 
LR1\_DR12\_VC05\_V70b & & & & & & & & \\ 
10 & 685,889.11 & 9,710.07 & & 573,240.49 & 3,869.13 && -- & -- \\ 
LR1\_DR12\_VC10\_V70a & & & & & & \\ 
10 & 547,246.29 & 9,727.10 & & 491,159.69 & 3,852.79 && -- & -- \\ 
LR1\_DR12\_VC10\_V70b & & & && & & & \\ 
10 & 667,435.32 & 10,885.22 & & 511,421.68 & 4,169.28 && -- & -- \\
\end{longtable}
\end{footnotesize}

\begin{footnotesize}
\setlength{\tabcolsep}{5pt}
\begin{longtable}[htbp]{lrrrrrrrr}
\toprule
& \multicolumn{2}{c}{\textbf{After BS}} && \multicolumn{2}{c}{\textbf{After LS}} && \multicolumn{2}{c}{\textbf{After ILS}}  \\ \cmidrule{2-3}\cmidrule{5-6}\cmidrule{8-9}
\multicolumn{1}{c}{\textbf{Instance}} & \multicolumn{1}{c}{\textbf{Cost}} & \multicolumn{1}{c}{\textbf{Time(s)}} && \multicolumn{1}{c}{\textbf{Cost}} & \multicolumn{1}{c}{\textbf{Time(s)}} && \multicolumn{1}{c}{\textbf{Cost}} & \multicolumn{1}{c}{\textbf{Time(s)}} \\ \midrule
\endhead
\bottomrule \caption{Results for instance with horizon = 360 (\textit{cont...})} \\
\endfoot
\bottomrule \caption{Results for instance with horizon = 360} \label{tab:resultados360}\\
\endlastfoot
LR1\_DR02\_VC01\_V6a & & & & & & & & \\ 
10 & 131,241.95 & 4.15 & & 117,805.53 & 4.87 & & 115,424.27 & 341.24 \\ 
100 & 110,554.82 & 64.40 & & 108,894.89 & 28.69 & & 108,543.07 & 347.44 \\ 
1000 & 109,040.08 & 525.90 & & 108,141.00 & 260.58 & & 108,141.00 & 342.42 \\ 
LR1\_DR02\_VC02\_V6a & & & & & & & & \\ 
10 & 302,309.98 & 5.68 & & 299,318.11 & 6.31 & & 298,176.81 & 573.27 \\ 
100 & 292,931.08 & 85.87 & & 291,667.79 & 42.60 & & 291,363.17 & 570.37 \\ 
1000 & 284,715.69 & 712.07 & & 283,788.36 & 315.11 & & 283,788.36 & 549.71 \\ 
LR1\_DR02\_VC03\_V7a & & & & & & & & \\ 
10 & 154,997.68 & 6.59 & & 147,784.22 & 5.96 & & 145,380.86 & 597.87 \\ 
100 & 135,588.49 & 99.77 & & 133,496.51 & 48.24 & & 133,092.34 & 587.45 \\ 
1000 & 128,179.60 & 834.49 & & 126,940.04 & 420.70 & & 126,940.04 & 551.24 \\ 
LR1\_DR02\_VC03\_V8a & & & & & & & & \\ 
10 & 177,638.23 & 5.70 & & 171,033.03 & 5.73 & & 167,193.45 & 449.40 \\ 
100 & 153,335.04 & 87.08 & & 149,730.10 & 37.00 & & 149,505.03 & 454.71 \\ 
1000 & 149,098.64 & 723.81 & & 146,883.21 & 302.77 & & 146,807.48 & 459.45 \\ 
LR1\_DR02\_VC04\_V8a & & & & & & & & 986.09 \\ 
10 & 149,713.33 & 10.83 & & 144,984.12 & 8.98 & & 143,917.35 & 976.18 \\ 
100 & 141,637.68 & 155.10 & & 140,249.09 & 79.94 & & 139,984.70 & 1,015.69 \\ 
1000 & 140,392.82 & 1,299.51 & & 139,266.26 & 766.34 & & 139,237.84 & 966.39 \\ 
LR1\_DR02\_VC05\_V8a & & & & & & & & 838.75 \\ 
10 & 126,186.66 & 8.93 & & 124,510.07 & 7.10 & & 123,656.41 & 791.69 \\ 
100 & 124,258.36 & 132.86 & & 123,178.11 & 67.60 & & 123,150.39 & 878.25 \\ 
1000 & 123,746.20 & 1,100.93 & & 122,859.78 & 601.20 & &  122,859.71 & 846.30 \\ 
LR1\_DR03\_VC03\_V10b & & & & & & & & \\ 
10 & 329,466.44 & 25.70 & & 318,285.87 & 29.13 & & 310,042.30 & 1,694.25 \\ 
100 & 289,974.00 & 276.36 & & 286,057.05 & 173.39 & & 285,561.93 & 1,625.33 \\ 
1000 & 264,144.66 & 2,525.70 & & 261,067.44 & 1,539.99 & & 260,807.85 & 1,563.02 \\ 
LR1\_DR03\_VC03\_V13b & & & & & & & & \\ 
10 & 352,033.91 & 71.14 & & 334,130.85 & 63.17 & & 326,090.47 & 3,411.63 \\ 
100 & 326,656.96 & 652.04 & & 321,280.99 & 555.78 & & 319,439.22 & 3,727.77 \\ 
1000 & 315,752.07 & 6,013.04 & & 312,857.51 & 4,307.37 & & 312,621.28 & 3,663.34 \\ 
LR1\_DR03\_VC03\_V16a & & & & & & & & \\ 
10 & 467,382.06 & 69.13 & & 454,372.57 & 72.68 & & 441,807.27 & 3,978.05 \\ 
100 & 390,382.71 & 544.77 & & 382,499.61 & 493.00 & & 380,730.40 & 3,808.74 \\ 
1000 & 356,252.63 & 5,333.36 & & 349,810.66 & 3,739.82 & & 349,216.16 & 3,743.24 \\ 
LR1\_DR04\_VC03\_V15a & & & & & & & & \\ 
10 & 305,977.18 & 131.15 & & 281,978.70 & 93.36 & & 264,208.59 & 6,613.50 \\ 
100 & 273,724.42 & 1,119.42 & & 262,976.69 & 1,171.03 & & 261,189.72 & 6,705.07 \\ 
1000 & 262,101.70 & 11,593.14 & & 257,127.67 & 9,716.80 & & 256,288.64 & 6,599.61 \\ 
LR1\_DR04\_VC03\_V15b & & & & & & & & \\ 
10 & 411,697.21 & 134.18 & & 377,519.63 & 118.47 & & 356,976.69 & 6,415.27 \\ 
100 & 365,132.58 & 1,185.99 & & 351,912.47 & 1,150.73 & & 349,552.52 & 6,526.01 \\ 
1000 & 347,431.99 & 12,248.41 & & 341,122.95 & 8,663.54 & & 340,710.34 & 6,540.66 \\ 
LR1\_DR04\_VC05\_V17a & & & & & & & & \\ 
10 & 266,056.76 & 181.00 & & 257,603.78 & 101.31 & & 255,334.59 & 9,778.30 \\
100 & 258,597.06 & 1,550.66 & & 255,262.39 & 1,243.77& & 254,692.16 & 10,599.58 \\ 
1000 & 256,888.26 & 17,771.13 & & 254,250.37 & 12,238.35 & & 254,040.64 & 10,406.66 \\ 
LR1\_DR04\_VC05\_V17b & & & & & & & & \\ 
10 & 383,190.49 & 180.94 & & 358,461.83 & 132.87 & & 345,029.81 & 9,102.59 \\ 
100 & 341,251.26 & 1,587.94 & & 330,007.26 & 1,227.64 & & 327,831.51 & 9,140.12 \\ 
1000 & 319,529.11 & 18,914.28 & & 312,507.44 & 8,669.68 & & 312,046.69 & 9,051.42 \\ 
LR1\_DR05\_VC05\_V25a & & & & & & & & \\ 
10 & 402,266.74 & 474.72 & & 388,250.32 & 146.55 & & 375,719.26 & 26,062.68 \\ 
100 & 396,689.52 & 4,426.28 & & 379,579.08 & 1,493.77 & & 373,871.33 & 25,895.11 \\ 
1000 & 386,148.10 & 63,455.80 & & & & \\ 
LR1\_DR05\_VC05\_V25b & & & & & & &  & \\ 
10 & 592,132.45 & 458.07 & & 573,194.47 & 194.42 & & 517,130.83 & 27,906.96 \\ 
100 & 561,152.88 & 4,212.32 & & 508,332.90 & 4,814.35 & & 494,627.59 & 26,204.78 \\ 
1000 & 497,030.22 & 68,865.93 & & & & \\ 
LR1\_DR08\_VC05\_V38a & & & & & & \\ 
10 & 1,087,526.69 & 3,385.03 & & 1,062,096.41 & 1,438.03 & & -- & -- \\ 
100 & 841,579.75 & 38,991.48 & & 683,046.79 & 42,092.79 & & -- & -- \\ 
LR1\_DR08\_VC05\_V40a & & & & & & \\ 
10 & 861,616.60 & 4,008.59 & & 706,365.61 & 7,023.42 & & -- & -- \\ 
100 & 694,011.53 & 42,493.51 & & 659,967.40 & 30,600.05 & & -- & -- \\ 
LR1\_DR08\_VC05\_V40b & & & & & & \\ 
10 & 1,030,976.88 & 3,270.36 & & 895,386.29 & 6,193.21 & & -- & -- \\ 
100 & 868,382.48 & 37,276.41 & & 804,362.42 & 43,404.17 & & -- & -- \\ 
LR1\_DR08\_VC10\_V40a & & & & & & \\ 
10 & 897,073.76 & 3,600.36 & & 746,645.15 & 6,872.98 & & -- & -- \\ 
100 & 702,624.40 & 42,222.05 & & 672,938.28 & 29,502.56 & & -- & -- \\ 
LR1\_DR08\_VC10\_V40b & & & & & & \\ 
10 & 1,032,161.48 & 3,278.19 & & 915,942.18 & 5,255.66 & & -- & -- \\ 
100 & 867,866.13 & 36,219.27 & & 794,420.68 & 40,328.66 & & -- & -- \\ 
LR1\_DR12\_VC05\_V70a & & & & & & \\ 
10 & 1,486,104.09 & 42,160.96 & & 1,193,627.62 & 30,094.65 & & -- & -- \\ 
LR1\_DR12\_VC05\_V70b & & & & & & \\ 
10 & 1,669,098.40 & 35,588.94 & & 1,435,397.61 & 41,286.04 & & -- & -- \\ 
LR1\_DR12\_VC10\_V70a & & & & & & \\ 
10 & 1,148,442.90 & 43,714.72 & & 1,046,678.91 & 26,082.87 & & -- & -- \\ 
LR1\_DR12\_VC10\_V70b & & & & & & \\ 
10 & 1,363,261.65 & 35,656.00 & & 1,306,455.89 & 22.382.99 & & -- & -- \\ 
\end{longtable}
\end{footnotesize}